\documentclass[runningheads]{llncs}

\usepackage{eccv}

\usepackage{eccvabbrv}

\usepackage{graphicx}
\usepackage{booktabs}
\usepackage{xcolor}
\usepackage{times}
\usepackage{enumitem}
\usepackage[table]{xcolor}
\usepackage{graphicx}
\usepackage{multirow}
\usepackage{multicol}

\usepackage[accsupp]{axessibility}  

\definecolor{first}{HTML}{fa9f7a}
\definecolor{second}{HTML}{fcbfa7}
\definecolor{third}{HTML}{fcd5c5}
\newcommand{\colorsquare}[1]{\colorbox{#1}{\hspace{6pt}\vphantom{A}}}

\usepackage{hyperref}

\usepackage{orcidlink}

\begin{document}

\title{NumColor: Enabling Numeric Color Generation in Text-to-Image Models} 
\title{Beyond Color Names: Precise Numeric Color Control in Text-to-Image Generation}
\title{Lost in Tokenization: Precise Numeric Color Control in Text-to-Image Generation}
\title{NumColor: Precise Numeric Color Control in Text-to-Image Generation} 
\titlerunning{NumColor}

\author{Muhammad Atif Butt\inst{1,2} \and Diego Hernández\inst{1,2} \and Alexandra Gomez-Villa\inst{1,2} \and Kai Wang\inst{3,1,4} \thanks{Corresponding author.} \and Javier Vazquez-Corral\inst{1,2} \and Joost Van De Weijer\inst{1,2}}

\authorrunning{Butt et al.}

\institute{Computer Vision Center, Spain \and
Computer Sciences Department, Universitat Autònoma de Barcelona, Spain \and
Program of Computer Science, City University of Hong Kong (Dongguan) \and
City University of Hong Kong
}

\maketitle

\begin{abstract}

Text-to-image diffusion models excel at generating images from natural language descriptions, yet fail to interpret numerical colors such as hex codes (\texttt{\#FF5733}) and RGB values (\texttt{rgb(255,87,51)}). This limitation stems from subword tokenization, which fragments color codes into semantically meaningless tokens that text encoders cannot map to coherent color representations. We present NumColor, that enables precise numerical color control across multiple diffusion architectures. NumColor comprises two components: a Color Token Aggregator that detects color specifications regardless of tokenization, and a ColorBook containing 6,707 learnable embeddings that map colors to embedding space of text encoder in perceptually uniform CIE Lab space. We introduce two auxiliary losses, directional alignment and interpolation consistency, to enforce geometric correspondence between Lab and embedding spaces, enabling smooth color interpolation. To train the ColorBook, we construct NumColor-Data, a synthetic dataset of 500K rendered images with unambiguous color-to-pixel correspondence, eliminating the annotation ambiguity inherent in photographic datasets. Although trained solely on FLUX, NumColor transfers zero-shot to SD3, SD3.5, PixArt-$\alpha$, and PixArt-$\Sigma$ without model-specific adaptation. NumColor improves numerical color accuracy by 4--9$\times$ across five models, while simultaneously improving color harmony scores by 10--30$\times$ on GenColorBench benchmark.

\end{abstract}

\section{Introduction}
\label{sec:intro}

Large-scale diffusion models have transformed text-to-image (T2I) generation, achieving remarkable visual fidelity. However, achieving precise color control remains a challenge. Although models reliably respond to coarse color descriptions such as basic color names (\eg, red), their performance degrades for precise color description. The degradation is especially severe when conditioned on numeric color codes (\eg, \textcolor[HTML]{E34234}{$\blacksquare$}~\texttt{\#E34234} in hexadecimal)~\cite{butt2025gencolorbench}. Because such codes are the standard for communicating color in professional design software (such as Adobe Suite, Canvas, and Blender), this limitation restricts the applicability of these models in advertising, brand-driven content creation, and product visualization, where strict color fidelity is required.

\begin{figure*}[t]
    \centering
    \includegraphics[width=\linewidth]{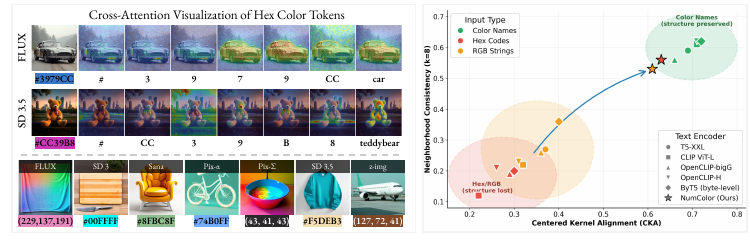}
    \vspace{-8mm}
    \caption{\textbf{Lack of Numeric Color Understanding.} \textit{Left:} Text encoders fragment hex codes into arbitrary subwords that attend to unrelated image regions, leading to incorrect color generation. \textit{Right:} CKA~\cite{kornblith2019similarity} and neighborhood consistency (k=8) between text embeddings and Lab space show that color names preserve perceptual structure, while hex codes and RGB values degrade by $\sim60\%$. NumColor recovers the structure comparable to color names.}
    \vspace{-5mm}
    \label{fig:motivation}
\end{figure*}

In Fig.~\ref{fig:motivation} (bottom-left), we provide 
representative failure cases in generation conditioned on numerical color codes for several models, including FLUX and SD3.  A recent benchmark of T2I models~\cite{butt2025gencolorbench} confirms these findings, reporting accuracy below 15\% on tasks involving numerical color specifications. This stems from how standard text encoders such as CLIP~\cite{radford2021learning} and T5~\cite{raffel2020exploring} tokenize these inputs~\cite{morovivCIC}: rather than treating them as coherent semantic units representing color, they are fragmented into separate tokens (\eg, \textcolor[HTML]{FF5733}{$\blacksquare$} \texttt{\#FF5733} becomes \texttt{\#}, \texttt{FF}, \texttt{57}, \texttt{33}), which not only lack any meaningful color representation in the embedding space, but also disrupt cross-attention. As shown in Fig.~\ref{fig:motivation} (top-left), each sub-token attends to arbitrary image regions rather than the target object, breaking color-object binding during generation. This is further supported by our embedding geometry analysis in Fig.~\ref{fig:motivation} (right) across five text encoders. While color names preserve strong consistency with the ground-truth color space, both RGB and hexadecimal codes exhibit an approximate 60\% degradation.


Beyond tokenization, the problem is further exacerbated by limited exposure to numerical color descriptions in the training data of T2I models. Image captions in large-scale text–image datasets predominantly describe colors using color names (e.g., “red,” “solferino,” “navy blue”), while explicit numerical color codes are rarely used. To address the lack of training data for numerical color supervision, we introduce \textbf{NumColor-Data}, a large-scale synthetic dataset. It contains 500,000 rendered images of diverse objects spanning a wide range of precisely specified colors. Each image provides unambiguous color-to-pixel correspondence, eliminating lighting and material ambiguities inherent in photographic datasets. 

In this work, we propose \textbf{NumColor}, a framework that enables precise numerical color generation in T2I diffusion models. While prior work such as ColorPeel~\cite{butt2025colorpeel} has explored color prompt learning, it is limited to a small fixed set of color names and requires retraining for each new color. NumColor instead addresses the problem in a rigorous fashion by proposing a new color encoder and a codebook that allow the models to understand precise color specifications. More in detail, we address the tokenizer fragmentation by introducing a tokenizer-agnostic Color Token Aggregator that identifies numerical color spans independently of the underlying text encoder. The detected colors are then mapped to the text embedding space via a  Color-based codebook, that we term \textit{ColorBook}: a compact set of learnable embeddings grounded in the perceptually uniform CIE Lab color space, injected directly into the text encoder to enable proper color-object binding during generation.

In summary, our main contributions are:
\vspace{-3mm}
\begin{itemize}
\item {\bf NumColor.} A framework for precise numerical color conditioning in T2I  models. It introduces a tokenizer-agnostic \textit{Color Token Aggregator} that operates independently of the tokenization scheme (e.g., CLIP or T5),  with \textit{ColorBook}, a learnable codebook that maps colors in the sRGB gamut to the text embedding space.
\item {\bf NumColor-Data.} The first large-scale dataset of objects with precisely numerical color descriptions. It contains over 500K images and covers the sRGB gamut, enabling clean supervision for numerical color conditioning. 
\item {\bf Results.} We evaluate NumColor on 9.8K prompts spanning 400+ colors from GenColorBench~\cite{butt2025gencolorbench}. Our method achieves 4--9$\times$ improvement in color accuracy and 10--30$\times$ improvement in color harmony across FLUX, SD3, SD3.5, PixArt-$\alpha$, and PixArt-$\Sigma$. The ColorBook, trained solely on FLUX, transfers zero-shot to SD3, SD3.5, PixArt-$\alpha$, and PixArt-$\Sigma$ without any model-specific adaptation.
\end{itemize}

\section{Related Work}
\label{sec:related_work}
\vspace{-2mm}
\subsection{Text-based Image Generation and Editing}

Alongside the rapid evolution of generative AI, remarkable advancements have been achieved in T2I generation and editing over the past several years. In particular, T2I diffusion models~\cite{ho2020ddpm,gu2022vector} have emerged as more effective paradigms, outperforming GANs~\cite{goodfellow2020generative}, VAEs~\cite{kingma2013vae}, autoregressive~\cite{esser2021taming}, and normalizing flows~\cite{dinh2014nice,dinh2016density} in T2I generation tasks. Diffusion models are probabilistic generative frameworks that learn underlying data distributions through iterative denoising processes, starting from Gaussian noise distribution. To enhance controllability, these models can be conditioned on various signals, including class labels~\cite{song2021ddim}, images~\cite{meng2022sdedit}, or text prompts~\cite{nichol2021glide}. With recent scaling of diffusion models, state-of-the-art T2I models such as SD3~\cite{esser2024scaling_sd3} and FLUX~\cite{flux2024} have significantly outperformed their predecessors~\cite{ramesh2022dalle2,chen2023pixartalpha,dai2023emu}.

However, T2I diffusion models lack inherent support for text-guided image manipulation. Prompt-to-Prompt~\cite{hertz2022prompt} pioneered controllable editing by aligning spatial layouts with text prompts via cross-attention, while InstructPix2Pix~\cite{brooks2022instructpix2pix} extended this through instruction tuning to enable text-guided manipulation. To circumvent the need for explicit image inversion, Imagic~\cite{kawar2022imagic} optimized text embeddings to align original image appearance with target semantic concepts, and Inversion-Free editing~\cite{xu2024inversion_free} reformulated the denoising process into a multi-step consistency framework to achieve stable virtual inversion. More recently, Flow Matching~\cite{lipman2022flow_matching,liu2022_rectified_flow} is integrated with large-scale transformers—driven by the Diffusion Transformer (DiT)~\cite{peebles2023scalable_dit} and FLUX~\cite{flux2024} have further enhanced image fidelity. Despite these notable advancements, most diffusion-based editing methods remain constrained by limited precise color control. While they excel at semantic alignment with textual prompts, they typically treat color as a global stylistic attribute rather than a spatially localized and continuous feature. This critical limitation motivates our work to develop a unified diffusion framework that integrates semantic localization and chromatic precision at the token level.

\vspace{-3mm}
\subsection{Diffusion-based Color Control}
With the rapid advancements in generative models~\cite{ramesh2022dalle2,ramesh2021zero,saharia2022imagen}, a variety of text-guided image editing approaches have been developed to enable controllable image modifications~\cite{hertz2023delta_DDS,meng2022sdedit,mokady2022null}. Meanwhile, unified models~\cite{deng2025emerging_bagel,wu2025omnigen2} integrate such editing capabilities through large-scale pretraining on massive paired datasets.
Another technical stream for achieving controllable generation is transfer learning applied to text-to-image (T2I) models~\cite{ruiz2022dreambooth,kumari2022customdiffusion}, which is also termed {T2I model adaptation} or {personalized generation}. This paradigm aims to adapt a pretrained T2I model to a {new concept} using user-provided images, binding the new concept to a unique token. Consequently, the adapted model can generate diverse renditions of the new concept under the guidance of text prompts. These adaptation methods can be categorized based on whether they fine-tune the entire T2I model~\cite{ruiz2022dreambooth,kumari2022customdiffusion} or freeze the T2I backbone while only optimizing auxiliary components~\cite{textual_inversion,wang2024mcti,dong2022dreamartist, yang2025echodistill}.
However, all these existing techniques struggle to achieve fine-grained control over color attributes in both image editing and generation tasks. To date, only a limited number of works have begun to address precise color generation~\cite{qin2025free_sadis,hou2025gencolor,qiu2025exploring_palette,tsai2025color_me_correctly}. For instance, Rich-Text~\cite{ge2023richtext} enhances color fidelity through multi-pass processing with global and local diffusion models, but this comes at the cost of high computational overhead and compromised color accuracy. In contrast, ColorPeel~\cite{butt2025colorpeel} introduces color prompt learning to improve the alignment between generated colors and user inputs. 
Nevertheless, this method requires extensive training and is constrained to handling only a small number of color names per training session.


Moreover, this problem persists in newer models. Recent work~\cite{butt2025gencolorbench} shows that state-of-the-art methods fail to reach even 10\% accuracy on numeric colors. We address this gap with NumColor to enable precise numerical color control in T2I diffusion models.


\vspace{-4mm}
\section{Method}
\label{sec:method}
\vspace{-3mm}

\begin{figure*}[t]
    \centering
    \includegraphics[width=\linewidth]{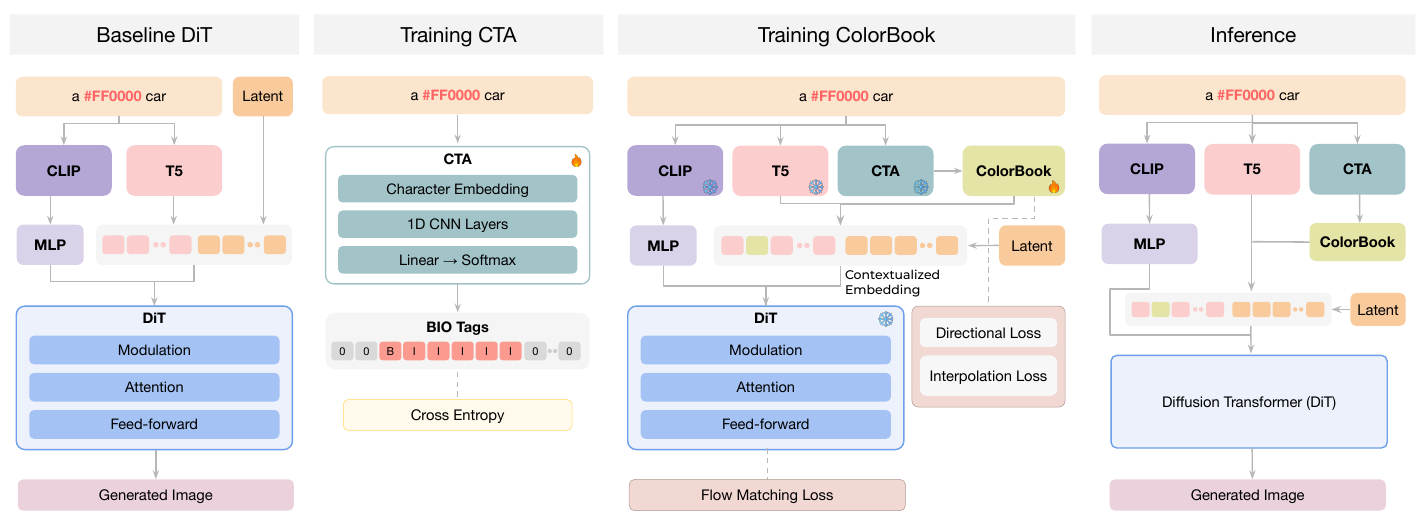}
    \vspace{-6mm}
    \caption{\textbf{Method overview.} \textit{(a) Baseline:} CLIP and T5 encode text; DiT generates images via iterative denoising. \textit{(b) Color Token Aggregator:} A character-level sequence labeler unifies fragmented color tokens using linear classifier, trained with cross-entropy on numeric color text prompts. \textit{(c) ColorBook Training:} Unified color tokens receive learned embeddings before T5 contextualization. The ColorBook (6,707 Lab anchors) is trained with flow matching loss on NumColor-Data, with directional and interpolation losses to preserve Lab geometry. Only ColorBook embeddings receive gradients; CLIP, T5, and DiT remain frozen.}
    \label{fig:method}
    \vspace{-6mm}
\end{figure*}

Given a text prompt containing a numeric color, our goal is to address two compounding failures: (i) tokenizer fragmentation, which prevents color codes from being treated as coherent semantic units, and (ii) the alignment gap between numerical color representations and the text encoder embedding space, leaving unified tokens perceptually ungrounded. To this end, we propose NumColor (Figure~\ref{fig:method}), a two-stage framework consisting of a Color Token Aggregator that resolves fragmentation in a tokenizer-robust manner, and ColorBook, which bridges the alignment gap by mapping numeric colors into the text embedding space through the perceptually uniform CIE Lab color space.

\vspace{-4mm}
\subsection{Preliminary}
\vspace{-2mm}
\noindent\textbf{Diffusion Transformers.} T2I diffusion models have evolved from U-Net based architectures to diffusion transformers which offer improved scalability and generation quality. Recent T2I models such as SD3 and Flux adopt this transformer based approach as a backbone. These models operate in VAE latent space, and learn to reverse a diffusion process that gradually adds noise to the images. Given a noisy latent $z_t$ at timestep $t$, the DiT is trained to predict the added noise (or velocity), enabling iterative denoising at inference. Modern DiTs employ rectified flow~\cite{esser2024scaling_sd3}, which learns straight-line trajectories between noise and data distributions, allowing for efficient few-step generation.

\noindent\textbf{Text Conditioning.} T2I diffusion transformers condition image generation on embeddings from pretrained text encoders, typically CLIP~\cite{radford2021learning} and T5~\cite{raffel2020exploring}. CLIP, trained through contrastive image-text alignment, provides semantically rich embeddings suited for visual concepts, while T5 offers better linguistic understanding for complex prompts. Current T2I models combine both to improve text conditioning. For instance, SD3 and SD3.5 employ CLIP-G and CLIP-L for pooled embeddings alongside T5-XXL, FLUX uses CLIP with T5-XXL, whereas PixArt uses T5 alone for text conditioning.

\vspace{-3mm}
\subsection{Color Token Aggregator}
\label{subsec:cta}
As demonstrated in Figure~\ref{fig:motivation}, subword tokenization fragments numerical color specifications into arbitrary tokens that lack color semantics.
The first challenge is identifying color tokens independent of how tokenizers fragment them. We introduce the Color Token Aggregator (CTA), a lightweight character-level module inspired by sequence labeling that detects numeric color specifications and unifies their fragments into single, semantically coherent tokens, regardless of the underlying vocabulary.
Given the token string from the tokenizer, each token is converted to a fixed-length character sequence (max 32 characters), embedded via a character embedding layer ($d_c = 64$), and processed by parallel 1D convolutions with kernel sizes $\{2, 3, 4\}$ to capture patterns at multiple scales. The convolution outputs pass through ReLU activations and max pooling, then concatenated to form a 256-dimensional token representation. A 4-layer transformer encoder with 4 attention heads contextualizes these representations across the token sequence. Finally, a linear classifier produces per-token emission scores for the three \{\texttt{B}, \texttt{I}, \texttt{O}\} (which represents Beginning, Inside, and Outside) tags:
\vspace{-2mm}
\begin{equation}
    \mathbf{s}_i = \mathbf{W}\mathbf{h}_i + \mathbf{b}, \quad \mathbf{s}_i \in \mathbb{R}^{3}
\vspace{-2mm}
\end{equation}

where $h_i \in \mathbb{R}^d$  is the contextualized representation of the i$-$th token from the transformer encoder, $\mathbf{W}$ is the projection matrix, and the output dimensions correspond to tag scores for \{\texttt{B}, \texttt{I}, \texttt{O}\}. A CRF layer~\cite{lafferty2001conditional} models tag transition dependencies, ensuring valid (Beginning, Inside, Outside) sequences. Training minimizes the negative log-likelihood under CRF. We curate 50K base prompts from DiffusionDB~\cite{wang2023diffusiondb} containing color name mentions, and replace color names with hex codes or RGB tuples. These prompts are tokenized across four text encoders (T5, CLIP-L/G, GPT-2), totaling 200K training samples. 
At inference, Viterbi decoding~\cite{shao2022viterbi} extracts optimal tag sequences, as color specifications are parsed into Lab coordinates for our ColorBook lookup.

\subsection{ColorBook}
\label{subsec:ColorBook}
\vspace{-2mm}
The CTA identifies numeric color fragmentation and unifies into a single color token, however these text tokens remain outside the learned vocabulary of the pretrained text encoder, since no pretrained embedding exists for the tokens like
\textcolor[HTML]{E34234}{$\blacksquare$}~\texttt{\#E34234} or \texttt{rgb(255,87,51)}. A naive approach to enable precise color control would be to learn a dedicated embedding for each color. However, the RGB color space contains over 16.7 million distinct values ($256^3$), making it infeasible to learn individual embeddings for every color. Even when transformed to CIE Lab space using standard sRGB transfer functions~\cite{fairchild2013color}, the sRGB gamut occupies a non-convex shaped region that remains prohibitively large for the embedding coverage. Learning this amount of color tokens directly would require prohibitive memory and training data.

\noindent\textbf{ColorBook construction.} We draw inspiration from vector quantization methods in generative modeling, where compact codebooks enable efficient representation of the high-dimensional space. In VQ-VAE~\cite{van2017neural}, a discrete codebook bridges encoder and decoder for image reconstruction. We repurpose this principle for color representation by constructing a codebook over color space itself, mapping perceptual color anchors to text encoder embeddings. Specifically, we define a codebook $\mathbf{E} \in \mathbb{R}^{K \times d}$ of anchor colors in the perceptually uniform CIE Lab space, which we refer to as \textbf{ColorBook}. We choose Lab over RGB because Euclidean distance in Lab correlates with perceived color difference, whereas RGB lacks this property. We uniformly sample Lab space at 5-unit intervals within sRGB gamut boundaries, yielding $K = 6{,}707$ anchor colors. This approach results in a reduction of over three orders of magnitude from naive per-color embeddings while preserving perceptual uniformity. Each anchor $\mathbf{a}_k \in \mathbb{R}^3$ maps to a learnable embedding $\mathbf{e}_k \in \mathbb{R}^{d}$, with $d = 4096$ to match T5 hidden dimensions.

\noindent\textbf{Embedding initialization.} Training embeddings from random initialization requires jointly learning perceptual color organization and alignment with the representation space of text encoder, which is a challenging optimization. Prior work demonstrates that pretrained initialization accelerates convergence, particularly with limited training data. We leverage this by initializing anchor embeddings from T5-XXL's color vocabulary, but rather than using individual color names (e.g., red, scarlet), which introduce linguistic biases, we use ISCC-NBS color category centroids~\cite{kelly1976color} that is a standard system of 267 perceptually-defined categories. For each category, we compute its LAB centroid, find the nearest anchor, and initialize its embedding from the T5 encoding of the category name; remaining anchors are initialized via distance-weighted interpolation from initialized neighbors. To handle continuous colors with a discrete ColorBook, we employ soft interpolation where for a query color $\mathbf{c} \in \mathbb{R}^3$, we retrieve the $k = 8$ nearest anchors $\mathcal{N}_k(\mathbf{c})$ and compute a weighted combination:

\begin{equation}
    \phi(\mathbf{c}) = \sum_{\mathbf{a}_j \in \mathcal{N}_k(\mathbf{c})} w_j \cdot \mathbf{e}_j, \quad w_j = \frac{\exp(-\|\mathbf{c} - \mathbf{a}_j\|_2 / \tau)}{\sum_{\mathbf{a}_i \in \mathcal{N}_k(\mathbf{c})} \exp(-\|\mathbf{c} - \mathbf{a}_i\|_2 / \tau)}
\end{equation}
where $\tau$ controls interpolation sharpness. We set $\tau = 2.0$, balancing sharp color assignment ($\tau \rightarrow 0$) against overly diffuse interpolation ($\tau \rightarrow \infty$). This enables smooth transitions across the color gamut while maintaining only 6,707 learnable embeddings.

\noindent{\textbf{Color embedding injection.}} Modern T2I diffusion models employ several text encoding strategies. Simpler architectures like PixArt use T5 only, while FLUX and SD3 combine T5 with CLIP-L/G encoders serving complementary roles: CLIP provides pooled embeddings summarizing entire prompt for global timestep conditioning, while T5 provides token-level sequence embeddings capturing fine-grained textual features. Since CLIP lacks per-token granularity, sequence embeddings of T5 become natural injection target. However, the precise injection point critically determines color-object association. Post-hoc replacement inserts color embeddings after T5 has contextualized the sequence, bypassing self-attention and preventing the model from binding colors to objects. We instead inject before transformer layers, replacing corresponding token embeddings with $\phi(\mathbf{c})$, then process the modified sequence. This pre-contextualization injection enables bidirectional attention between color and object tokens (e.g.,
\texttt{\#FF5733} $\leftrightarrow$ car), facilitating color-object association during generation (Figure~\ref{fig:cross-attn}).

\vspace{-5mm}
\subsection{Dataset Curation}
\vspace{-2mm}
Training a ColorBook requires image-text prompt pairs where the specified color exactly matches the rendered appearance of the objects. Existing large-scale datasets fail this requirement. For instance, Butt et. al~\cite{butt2025gencolorbench} show that LAION~\cite{schuhmann2022laionb} captions rarely specify precise colors, and when they do, the photographed objects exhibit specular highlights, subsurface scattering, ambient occlusion, and environmental color casts that confound the mapping between prompt color and pixel color. A model trained on such data may perform well on coarse basic color generation, however, it would remain challenging to disentangle color specification from material and lighting effects.

\begin{figure*}[t]
    \centering
    \includegraphics[width=\linewidth]{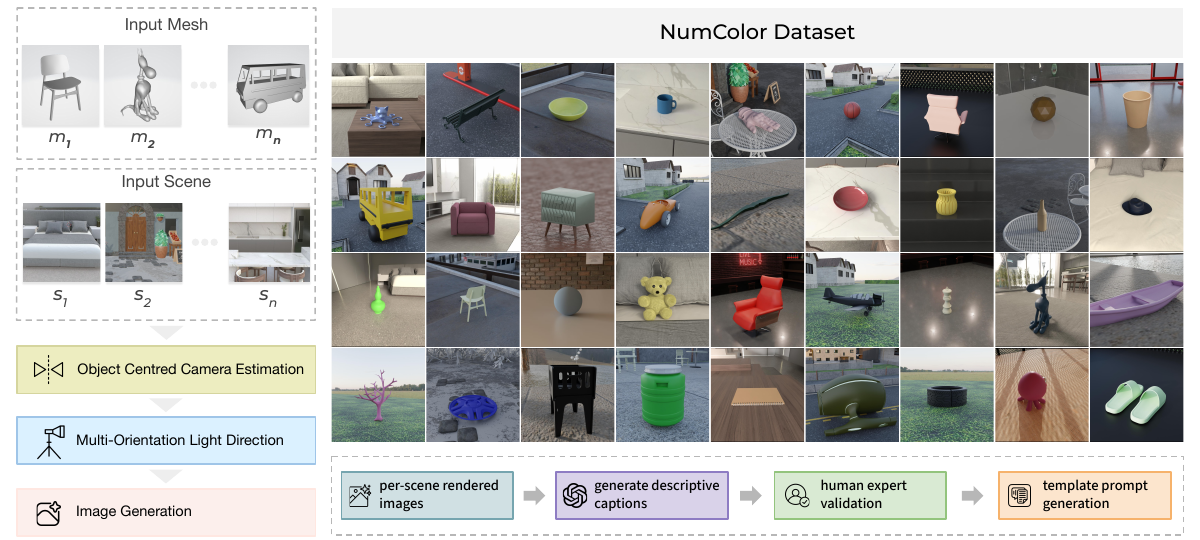}
    \caption{\textbf{NumColor-Data generation pipeline.} We render 3D meshes from Objaverse-XL in diverse indoor and outdoor scenes. Each object is assigned a uniform Lambertian material with albedo set to the target Lab color. We use object-centered camera estimation and multi-orientation lighting to ensure diverse viewpoints. The dataset comprises 500K images spanning varied objects, scenes, and colors from ColorBook anchors. \textbf{Caption generation}---rendered images are paired with descriptive captions, validated by human experts, and converted to template prompts.}
    \label{fig:dataset}
    \vspace{-6mm}
\end{figure*}

To address this shortcoming, we construct \textbf{NumColor-Data}, a synthetic dataset providing unambiguous color supervision, as shown in Figure~\ref{fig:dataset}. We source 100 3D meshes from the Objaverse-XL~\cite{deitke2023objaverse} dataset. These meshes cover broad categories of objects, including furniture, vehicles, and items of daily use. Further, we collect 34 indoor and outdoor scenes from BlenderKit~\cite{blenderkit}, and the HDRIs from~\cite{Haven} to create a diverse dataset. Each mesh is assigned a uniform Lambertian material with albedo set to a target Lab color. We render using Blender Cycles under the D65 illuminant with controlled intensity to ensure accurate color reproduction. The prompt templates are listed in the supplementary material. The final dataset comprises 500K images. For each object-scene pair, we sample the colors from the 6,707 Lab anchors, ensuring coverage throughout the color space. We use a 90/10 train/validation split stratified by objects. The rendered albedo exactly matches the prompt specification, providing direct supervision for learning the ColorBook without confounding material properties or lighting. 

\vspace{-3mm}
\subsection{Training Objective}
\vspace{-2mm}
Learning color embeddings requires both accurate numeric color token unification and alignment between the learned embedding space and color structure. We achieve these objectives through a two-stage training.

\noindent \textbf{Stage 1}. We train the CTA aggregator on 200K text prompts (see Section~\ref{subsec:cta} for details) to identify numeric color tokens from tokenized strings. The encoder is trained on 200K synthetic text prompts using negative log-likelihood under the CRF: 
\begin{equation}
    \mathcal{L}_{\text{CTA}} = -\log P(\mathbf{y}^* | \mathbf{p}),    
\end{equation}
where $\mathbf{p}$ is the tokenizer prompt string, and $\mathbf{y}^*$  is the ground-truth encoded sequence of the prompt. The CRF layer enforces valid tag transitions (such as, \texttt{I} cannot follow \texttt{O}) which improves the sequence coherence over independent token classification.

\noindent \textbf{Stage 2}. We freeze the CTA aggregator, and we train the ColorBook embeddings $\mathbf{E}$ on NumColor-Data. The base diffusion model and T5~\cite{raffel2020exploring} text encoder remain frozen; only $\mathbf{E}$ receives gradients. For rectified flow models, the primary objective is flow-matching:
\begin{equation}
    \mathcal{L}_{\text{flow}} = \mathbb{E}_{t, z_0, z_1}\left[\|v_\theta(z_t, t, \mathcal{E}(\mathbf{p}; \phi)) - (z_1 - z_0)\|_2^2\right],
\end{equation}
where $z_t = (1-t)z_0 + tz_1$, and $\mathcal{E}(\mathbf{p}; \phi)$ denotes T5 encoding~\cite{raffel2020exploring} with color embeddings injected via $\phi$. The flow matching objective trains the ColorBook to produce embeddings that generate correct colors, but provides no explicit supervision on the \emph{geometry} of the embedding space. Without additional constraints, perceptually similar colors may map to distant embeddings, degrading interpolation quality and generalization to unseen colors. Therefore, We introduce two auxiliary losses that enforce correspondence between Lab color space and embedding geometry. 

\textit{Directional alignment} encourages embedding displacements to reflect perceptual displacements in Lab space:
\begin{equation}
    \mathcal{L}_{\text{dir}} = \mathbb{E}_{\mathbf{c}_i, \mathbf{c}_j}\left[1 - \cos\left(\phi(\mathbf{c}_i) - \phi(\mathbf{c}_j),\, \psi(\mathbf{c}_i - \mathbf{c}_j)\right)\right]
\end{equation}

where $\mathbf{c}_i , \mathbf{c}_j \in \mathbb{R}^3$ are Lab coordinates sampled from anchor set, $\phi: \mathbb{R}^3 \rightarrow \mathbb{R}^d$ denotes the ColorBook mapping from Lab space to text embedding space, and $\psi: \mathbb{R}^3 \rightarrow \mathbb{R}^d$ is a learned linear projection mapping Lab differences to embedding dimension.

\textit{Interpolation consistency} enforces that perceptual midpoints in Lab space map to embedding midpoints, ensuring smooth transitions:
\begin{equation}
    \mathcal{L}_{\text{interp}} = \mathbb{E}{(\mathbf{c}_i, \mathbf{c}_j) \sim \mathcal{P}}\left[\left|\phi\!\left(\tfrac{\mathbf{c}_i + \mathbf{c}_j}{2}\right) - \tfrac{1}{2}\bigl(\phi(\mathbf{c}_i) + \phi(\mathbf{c}_j)\bigr)\right|_2^2\right]
\end{equation}

where $\mathbf{c}_i , \mathbf{c}_j$ are Lab color coordinates sampled from the anchor set and $\mathcal{P}$ denotes the distribution over color pairs. The left term computes the embedding of the Lab midpoint, while the right term computes the midpoint of two embeddings; minimizing their difference enforces linearity in the learned mapping. 

The total objective combines reconstruction with geometric regularization:
\begin{equation}
    \mathcal{L} = \mathcal{L}_{\text{flow}} + \lambda_d.\mathcal{L}_{\text{dir}} + \lambda_i.\mathcal{L}_{\text{interp}}
    \vspace{-3mm}
\end{equation}
where $\lambda_d$ and $\lambda_i$ are hyperparameters to control relative contribution of regularizers.
\vspace{-3mm}
\section{Experiments}
\vspace{-2mm}
\label{sec:experiments}
\subsection{Experiment Setup}
\subsubsection{Implementation Details.}
We use FLUX.1-dev~\cite{flux2024} as the primary backbone. In stage 1, the CTA is trained for 50K steps using AdamW~\cite{loshchilov2018decoupled} with learning rate $1\times 10^{-4}$, batch size 128, and weight decay 0.01. In stage 2, the FLUX transformer and T5 encoder are frozen; only ColorBook embeddings $\mathbf{E}$ $\in$ $\mathbb{R}^{6707 \times 4096}$ are learned. ColorBook is trained for 100K steps with learning rate $1\times 10^{-5}$ and batch size 16 on $4\times$ A100 64GB GPUs.
\vspace{-10mm}
\subsubsection{Baselines.} We evaluate against state-of-the-art T2I diffusion models spanning diverse architectures: FLUX.1-dev~\cite{flux2024}, SD3 and SD3.5~\cite{esser2024scaling_sd3}, PixArt-$\alpha$ ~\cite{chen2023pixartalpha}, PixArt-$\Sigma$~\cite{chen2024pixartsigma}, Sana~\cite{xie2024sana}, and CogView4~\cite{ding2021cogview}. These baselines represent both single-encoder and multi-encoder architectures, enabling analysis of how different text conditioning strategies handle numeric color specifications. We integrate NumColor with FLUX SD3, SD3.5, PixArt-$\alpha$, and PixArt-$\Sigma$ to evaluate both trained performance and zero-shot transfer.
\vspace{-6mm}
\subsubsection{Evaluation Metrics.} We adopt the evaluation protocol from GenColorBench~\cite{butt2025gencolorbench} benchmark, which measures color accuracy using $\Delta E$ (CIE2000)---the perceptual color distance between specified and generated colors. Lower $\Delta E$ indicates better color fidelity, with $\Delta E < 1$ imperceptible to humans and $\Delta E < 5$ considered acceptable. We evaluate on three color granularities including ISCC-NBS Level 1, ISCC-NBS Level 3, and CSS3/X11 web colors. We use 9.8K numerical color prompts from GenColorBench covering 400+ colors, which evaluate hex code and RGB triples. Additionally, we compute color harmony metrics to assess the aesthetic coherence in the generated images.
\vspace{-8mm}
\subsection{Qualitative Results}

\begin{figure}[!t]
    \centering
    \includegraphics[width=\linewidth]{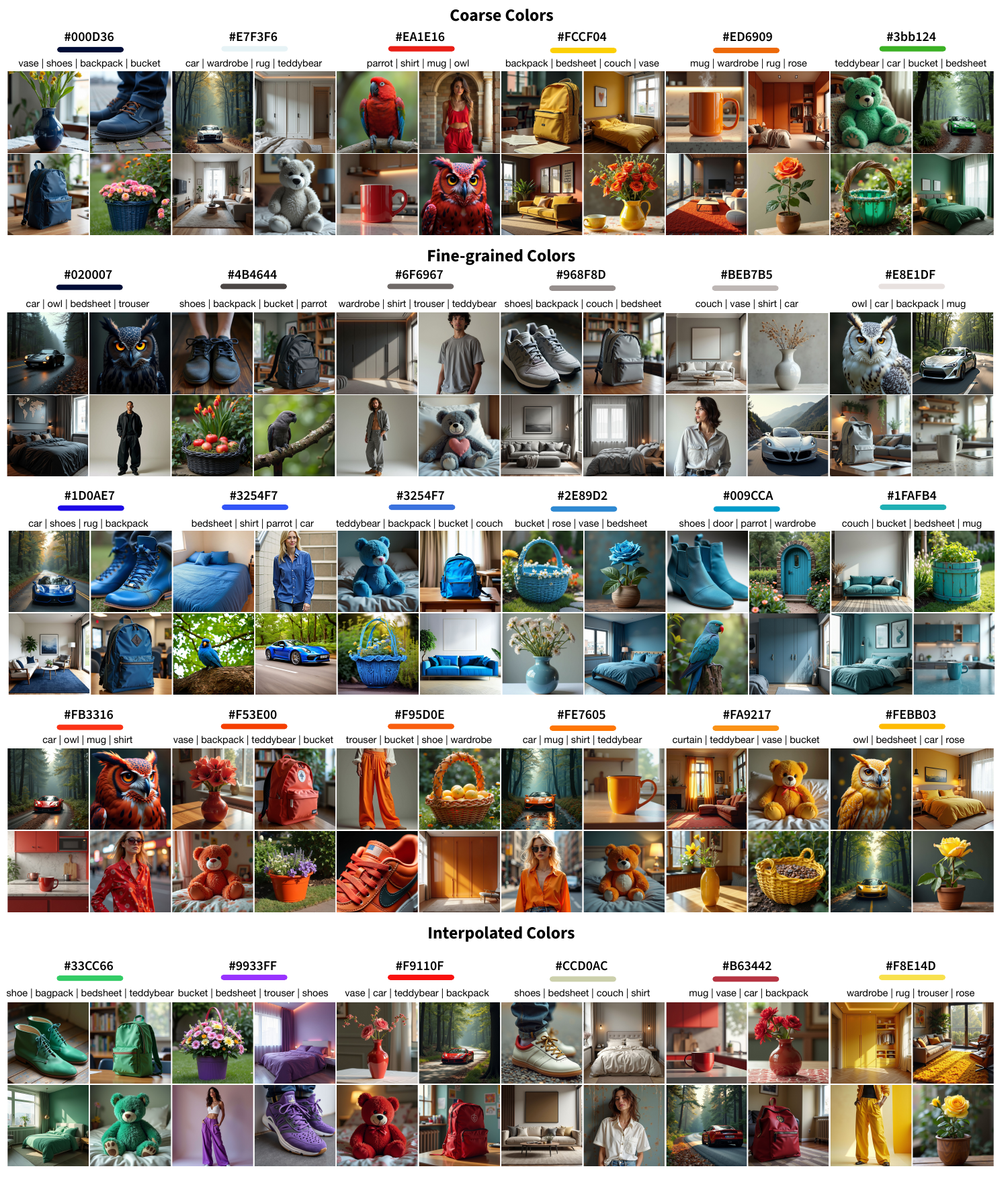}
    \vspace{-8mm}
    \caption{Qualitative results on FLUX with NumColor. We evaluate \textit{coarse colors} spanning primary hues from deep navy to saturated green; \textit{fine-grained colors} including perceptually adjacent grayscale, blue-to-cyan, and red-to-yellow colors; and \textit{interpolated colors} synthesized via interpolating ColorBook anchors. NumColor maintains accurate color reproduction across objects.}
    \vspace{-8mm}
    \label{fig:qual_res}
\end{figure}

We present qualitative results demonstrating color control across three dimensions: (i) coarse colors spanning the primary hue spectrum, (ii) fine-grained colors requiring perceptual distinction, and (iii) interpolated colors---which are not from NumColor anchors. Figure~\ref{fig:qual_res} shows generations using our method on FLUX that accurately reproduce the primary colors across semantically diverse objects, confirming that the color embeddings transfer independently of object category. We also demonstrate fine-grained color transitions, including grayscale---from black via gray to white where adjacent swatches differ by $\Delta E < 15$. Similarly, we show the color transition from blue to cyan, and red to yellow. Despite the perceptual subtlety of these distinctions, our method maintains clear differentiation across the color progression, demonstrating that NumColor preserves fine color structure in embedding space. Lastly, we demonstrate color generation for colors that are not available as anchors in NumColor. These colors are synthesized using interpolation looking from neighboring anchors as described in Section~\ref{subsec:ColorBook}. Accurate reproduction of these out-of-sample colors confirms that our learned embeddings are from a continuous, interpolable manifold rather than a discrete lookup table.

\begin{figure}[t]
    \centering
    \includegraphics[width=\linewidth]{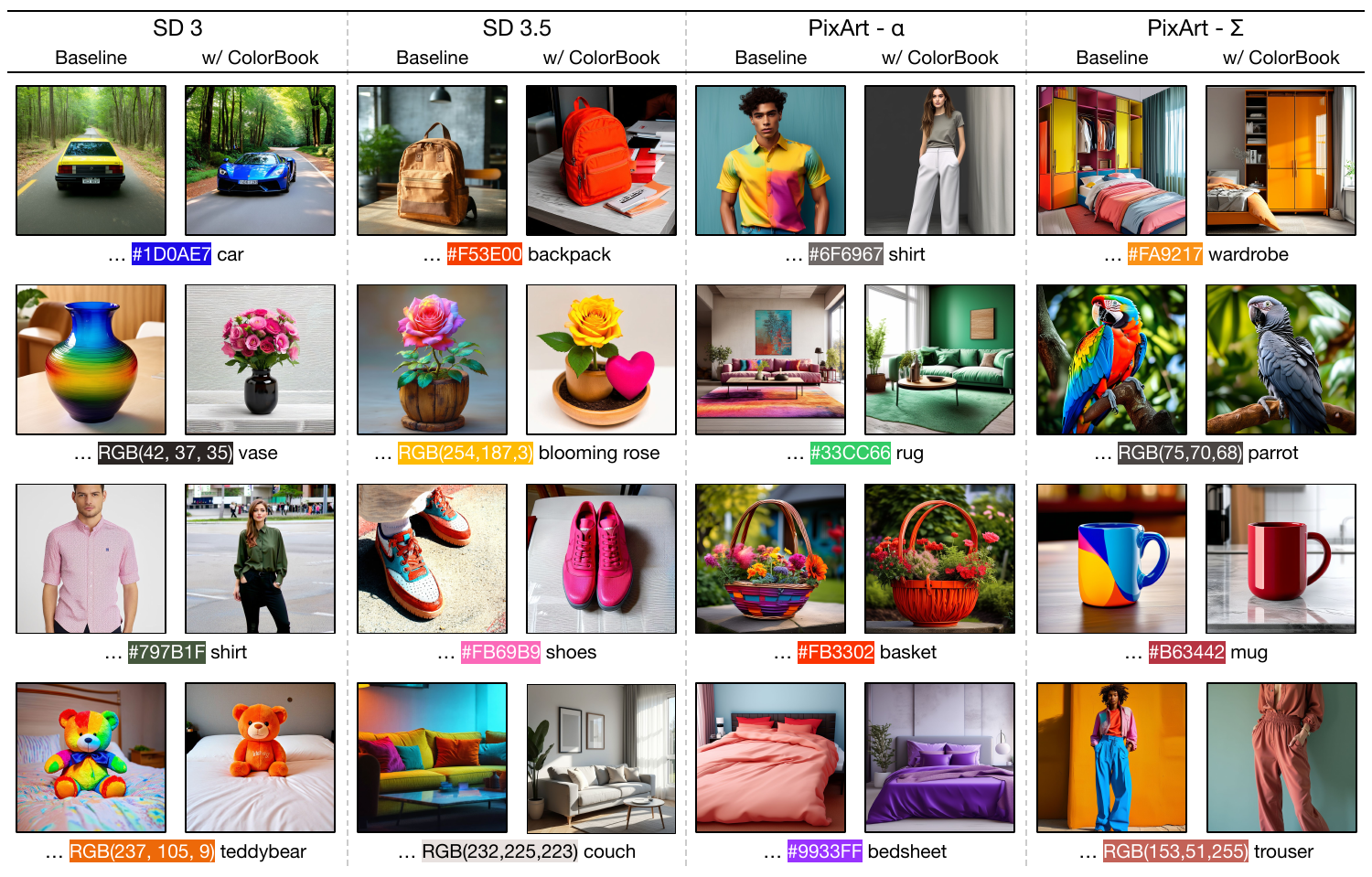}
    \vspace{-6mm}
    \caption{\textbf{Cross-model generalization.} Baseline models versus NumColor integration across SD3, SD3.5, PixArt-$\alpha$, and PixArt-$\Sigma$. NumColor, trained on FLUX, transfers zero-shot to other diffusion models. Baselines exhibit incorrect hue mapping, ignored specifications, and rainbow artifacts; NumColor resolves these while preserving image quality.}
    \label{fig:generalization}
    \vspace{-6mm}
\end{figure}

In Figure~\ref{fig:generalization}, we demonstrate cross-model transfer by comparing baselines with NumColor-integrated models. Although primarily trained on FLUX, NumColor transfers zero-shot to SD3, SD3.5, PixArt-$\alpha$, and PixArt-$\Sigma$ through their shared T5 encoder. Baseline models exhibit characteristic failures. For instance, SD3 renders \textcolor[HTML]{1D0AE7}{$\blacksquare$}\texttt{\#1D0AE7} as yellow; SD3.5 ignores \textcolor[HTML]{F53E00}{$\blacksquare$}\texttt{\#F53E00} entirely, producing a beige backpack. 
With NumColor, all models correctly reproduce specified colors while maintaining generation quality. Challenging cases including near-neutral tones and desaturated hues are handled consistently across architectures. Further results are in supplementary.

\vspace{-4mm}
\subsection{Embedding Space Analysis}
\begin{figure}[t]
    \centering
    \includegraphics[width=\linewidth]{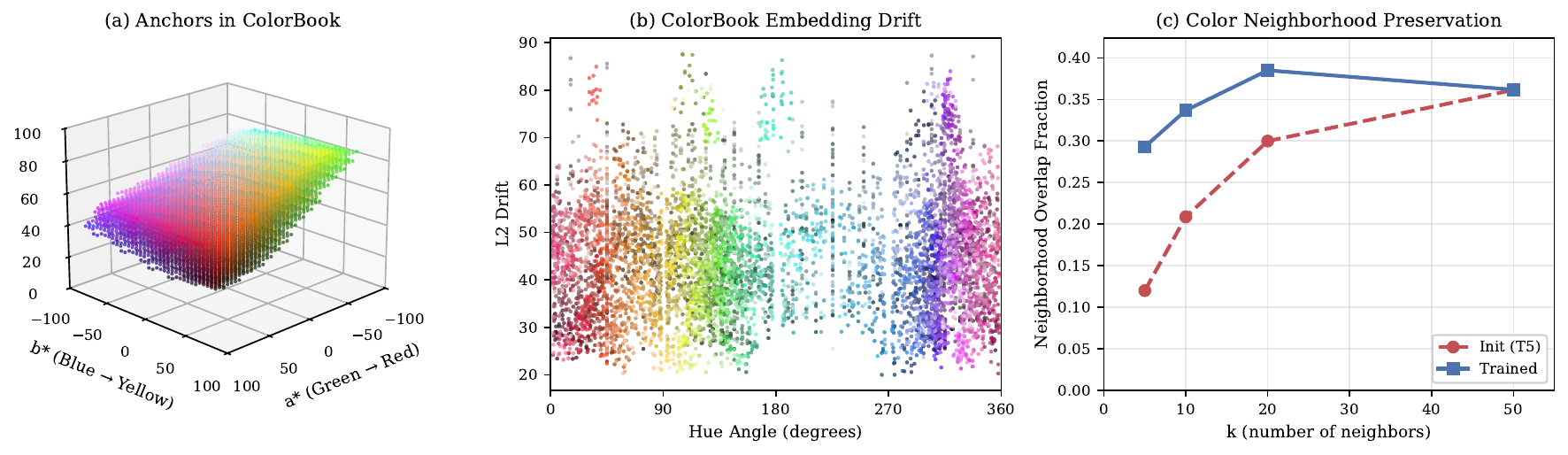}
    \vspace{-6mm}
    \caption{Embedding space analysis of NumColor. (a) Distribution of 6,707 color anchors in CIE Lab space with 5-unit spacing. (b) L2 drift between initial (T5-encoded) and trained embeddings versus hue angle, showing uniform adaptation across the color spectrum. (c) Nearest Neighbor overlap between LAB and embedding space. Initial T5 embeddings show high consistency at small k due to shared category names but degrade at larger k; trained embeddings maintain preservation at larger neighborhoods, reflecting perceptual structure beyond linguistic clustering.}
    \label{fig:drift_analysis}
\end{figure}

Beyond evaluating generation quality, we analyze the geometric properties of the learned ColorBook to verify that training produces perceptually meaningful representations rather than arbitrary embeddings that happen to reconstruct colors (Fig.~\ref{fig:drift_analysis}). We measure the $L_2$ drift between initial T5-encoded embeddings and their trained counterparts as a function of hue angle (Fig.~\ref{fig:drift_analysis}b). The drift distributes uniformly across the hue spectrum, indicating that training adapt to all color regions equally rather than overfitting to specific hues. We also observe the neighborhood preservation analysis (Figure~\ref{fig:drift_analysis}c), which quantifies whether perceptual neighbors in Lab space remain neighbors in embedding space. Initial T5 embeddings exhibit high nearest neighbors overlap at small k, however this is an artifact of our ISCC-NBS initialization as colors within the same category share identical embeddings. Trained embeddings exhibit moderate overlap at small k while sustaining preservation at larger neighborhoods, confirming that training reorganizes the embedding space to reflect perceptual similarity rather than linguistic categorization. Moreover, learned geometric organization enables NumColor to generalize to colors unseen during training through smooth interpolation between anchors. This improvement is also reflected in Fig.~\ref{fig:motivation} (right), where NumColor embeddings (stars) align with the Lab color structure approaches the consistency observed for color names.

\vspace{-3mm}
\subsection{Quantitative Comparisons}

\begin{table}[t]
\caption{Numerical Color Understanding performance of T2I models on GenColorBench. Results across three color specification granularities.
\colorsquare{first} \colorsquare{second} \colorsquare{third} indicate top-3 performers.}
\centering
\vspace{-3mm}
\resizebox{0.85\linewidth}{!}{
\begin{tabular}{lcc|ccc|cc}
\hline
& & & \multicolumn{3}{c|}{GenColorBench Metrics $\uparrow$} & \multicolumn{2}{c}{Color Harmony $\downarrow$} \\
Model & Resolution & Type & ISCC-L1 & ISCC-L3 & CSS3/X11 & RGB & Hex \\
\hline
FLUX & 1024 & DM & 13.82 & 7.74 & 5.86 & 247.4 & 134.9\\
Sana & 1024 & DM & 25.10 & 12.85 & 9.45  &  643.1 & 105.5 \\
SD3.5 & 1024 & DM & 14.92 & 7.53 & 5.78  &  518.3 & 117.3 \\
Pixart-$\alpha$ & 1024 & DM & 9.45 & 5.42 & 4.21  & 638.9 & 366.1 \\
SD3 & 1024 & DM & 11.80 & 5.92 & 4.63  & 826.4 & 251.4 \\
Pixart-$\Sigma$ & 1024 & DM & 10.21 & 5.11 & 4.09 & 673.7 & 673.7 \\
CogView4 & 1024 & DM & 10.95 & 5.32 & 4.25  & 476.9 & 253.0 \\
\hline
FLUX w/ NumColor & 1024 & DM         & \cellcolor[HTML]{fa9f7a}55.71 & \cellcolor[HTML]{fa9f7a}48.02 & \cellcolor[HTML]{fa9f7a}51.96 & \cellcolor[HTML]{fa9f7a}16.90 & \cellcolor[HTML]{fa9f7a}18.47 \\
SD 3 w/ NumColor & 1024 & DM         & \cellcolor[HTML]{fcd5c5}49.23 & \cellcolor[HTML]{fcd5c5}39.46 & \cellcolor[HTML]{fcbfa7}47.19  & \cellcolor[HTML]{fcd5c5}22.64 & \cellcolor[HTML]{fcd5c5}22.35 \\
SD3.5 w/ NumColor & 1024 & DM       & \cellcolor[HTML]{fcbfa7}51.72 & \cellcolor[HTML]{fcbfa7}43.35 & \cellcolor[HTML]{fcd5c5}46.87  & 24.38 & 25.51\\
PixArt-$\alpha$ w/ NumColor & 1024 & DM & 44.93 & 36.17 & 38.51  & \cellcolor[HTML]{fcbfa7}19.54 & \cellcolor[HTML]{fcbfa7}19.87 \\
PixArt-$\Sigma$ w/ NumColor & 1024 & DM & 48.56 & 40.83 & 43.90  & 24.08 &  25.79 \\
\hline
\end{tabular}
}
\label{tab1:results}
\vspace{-5mm}
\end{table}

Table~\ref{tab1:results} presents numerical color understanding performance on GenColorBench benchmark across three granularities. We evaluate all models on 9.8K prompts using both hex and RGB formats. Baseline models fail regardless of granularity: FLUX achieves only 13.82\% at ISCC-L1 despite just 13 coarse categories—marginally above the 7.7\% random baseline—degrading to 7.74\% at ISCC-L3 and 5.86\% at CSS3/X11. This pattern persists across architectures, confirming that the bottleneck is not color discrimination but text encoders' inability to extract meaning from tokenized color specifications. Whereas, NumColor integration yields substantial improvements across all models and granularities. On FLUX, accuracy increases from 13.82\% to 55.71\% at ISCC-L1, 7.74\% to 48.02\% at ISCC-L3, and 5.86\% to 51.96\% at CSS3/X11—representing 4--9$\times$ relative improvement. Gains are largest at fine granularities where baselines approach random chance, demonstrating that learned embeddings preserve perceptual distinctions in Lab space. Cross-model transfer validates universality: using NumColor trained exclusively on FLUX, we observe average improvements of +37.4\% on SD3, +36.8\% on SD3.5, +33.5\% on PixArt-$\alpha$, and +37.3\% on PixArt-$\Sigma$.

We additionally evaluate color harmony following Cohen-Or et al.~\cite{cohen2006color}, which measures conformance to harmonic templates on the hue wheel. Baseline models exhibit poor harmony scores such as 247--826 for RGB, and 105--673 for hex, as incorrect color reproduction often yields incoherent palettes. NumColor reduces these scores to 16--24 for RGB and 18--26 for hex which yields a 10--30$\times$ improvement demonstrating that precise color control simultaneously enhances aesthetic coherence.

\begin{figure}[t]
    \centering
    \includegraphics[width=0.95\linewidth]{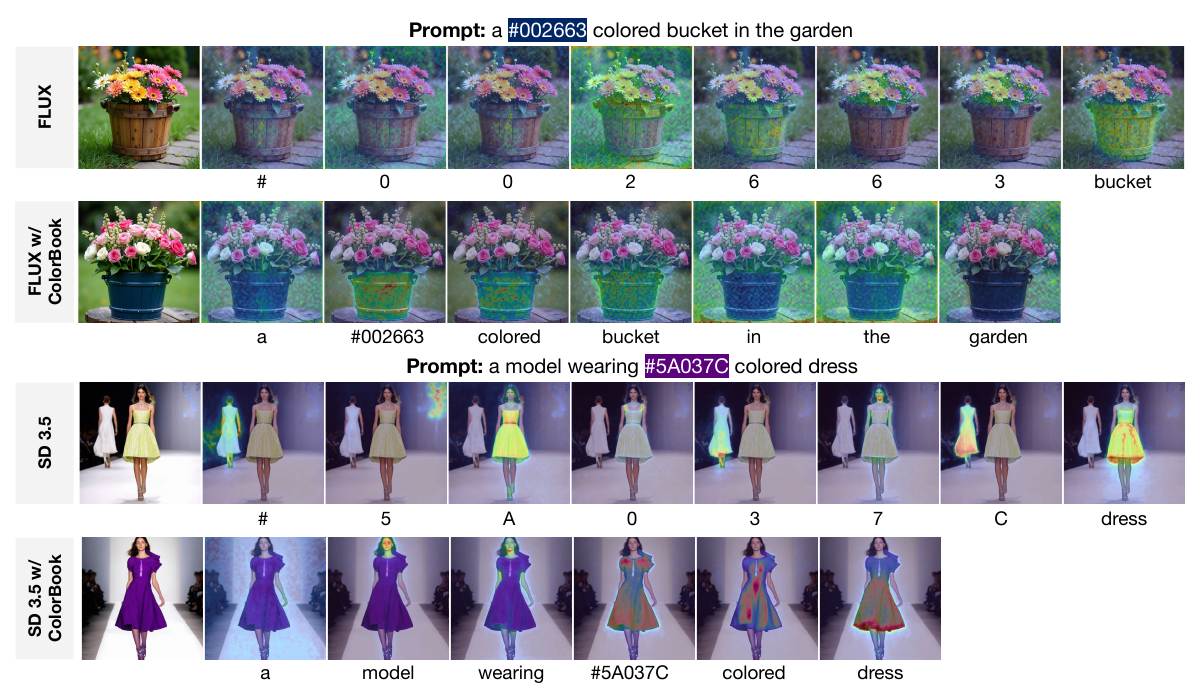}
    \vspace{-3mm}
    \caption{Cross-attention maps for hex color specifications. Baseline models fragment hex codes into character tokens with scattered attention. NumColor processes colors as unified embeddings, focusing attention on target objects and enabling accurate color control.}
    \vspace{-4mm}
    \label{fig:cross-attn}
\end{figure}

\vspace{-6mm}
\subsection{Ablations}
\begin{figure}[t]
    \centering
    \includegraphics[width=0.95\linewidth]{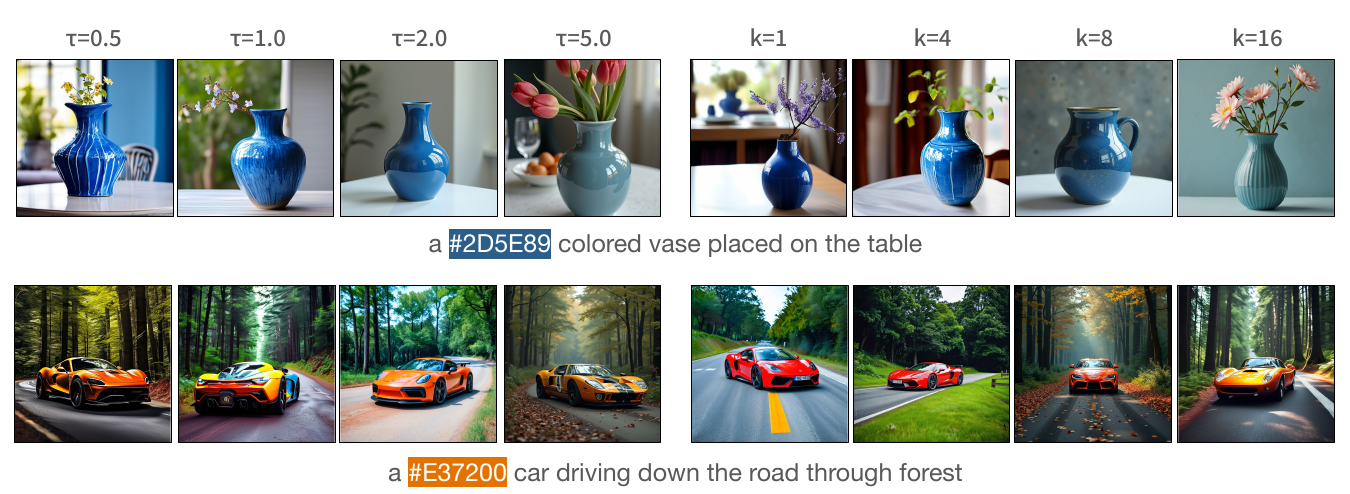}
    \vspace{-3mm}
    \caption{Hyperparameter ablation. Low k causes hard quantization artifacts. High $\tau$ leads to color desaturation. $\tau=2.0$, $k=8$ balance accurate color reproduction with smooth interpolation.}
    \label{fig:ablation}
    \vspace{-4mm}
\end{figure}

\vspace{-2mm}
We conduct ablation studies to validate design choices, examining cross-attention behavior after NumColor integration and hyperparameter sensitivity. Figure~\ref{fig:cross-attn} visualizes cross-attention maps comparing baseline models with NumColor. Without NumColor, both FLUX and SD 3.5 fragment hex codes into individual character tokens (Figure \ref{fig:motivation}), resulting in scattered attention patterns across unrelated image regions. With NumColor, color specification is processed as a unified semantic unit, concentrating attention on the target object. This consolidation enables precise color-to-object binding. Further, we analyze  effect of softmax temperature $\tau$ and color neighborhood size on the color generation, shown in Figure~\ref{fig:ablation}. The temperature parameter controls the sharpness of interpolation weights over k-nearest anchors. At low $\tau$, weights concentrate on closest anchors, while high $\tau$ distributes weights more uniformly, causing color desaturation. At k=1, hard assignment to the nearest anchor produces quantization artifacts. We choose $\tau$=2.0 and k=8, balancing accurate color reproduction with smooth interpolation.

\begin{figure}[t]
    \centering
    \includegraphics[width=0.70\linewidth]{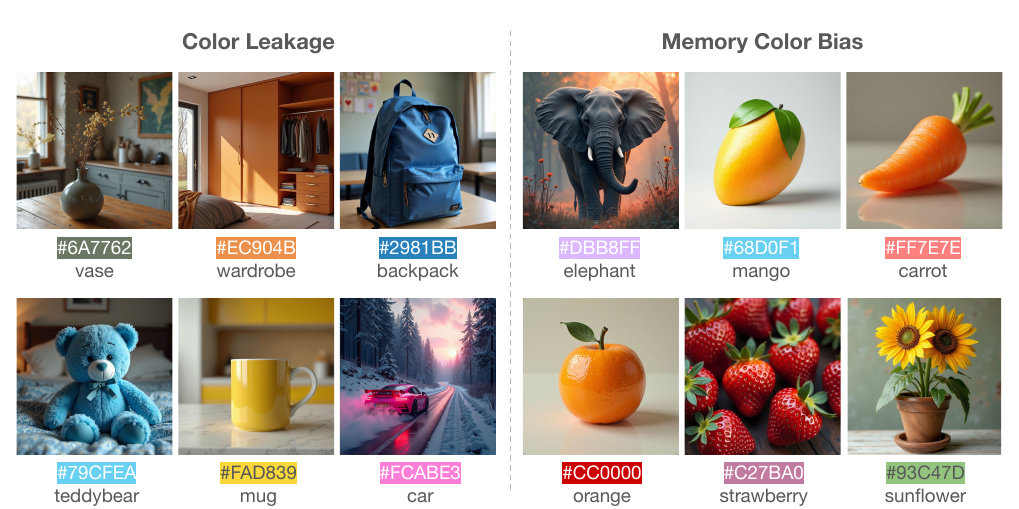}
    \vspace{-3mm}
    \caption{\textbf{Failure cases.} \textit{Color Leakage:} Specified color appears on the object but bleeds into surroundings due to lack of spatial grounding. \textit{Memory Color Bias:} Objects with canonical binding resist recoloring, as pretrained model overrides the embedding, defaulting to memorized colors.}
    \label{fig:limitations}
    \vspace{-6mm}
\end{figure}

\section{Limitations}
While our method enables precise numerical color control, we identify two failure modes inherent to underlying diffusion models (Figure~\ref{fig:limitations}). \textit{Color leakage} occurs when the specified color bleeds into surrounding elements as cross-attention lacks explicit spatial grounding to enforce boundaries. \textit{Memory color bias} causes objects with strong canonical associations to resist recoloring; despite specifying green for sunflowers, the model defaults to memorized yellow, reflecting object-color priors that override the color embedding. Both limitations represent fundamental challenges in controlling pretrained generative models, not specific to NumColor.

\vspace{-4mm}
\section{Conclusion}
\vspace{-3mm}

We presented NumColor, a plug-and-play framework for precise numeric color control in T2I generation. NumColor addresses two compounding failures: tokenizer fragmentation via a tokenizer-agnostic Color Token Aggregator, and the alignment gap between color specifications and embedding space via ColorBook operating in perceptually uniform CIE LAB space. Both components require no adaptation to pretrained diffusion models. To enable training, we proposed NumColor-Data, a synthetic dataset of 500K images with unambiguous color-to-pixel correspondence. The trained ColorBook transfers zero-shot to multiple diffusion models, achieving $4-9×\times$ improvement in color accuracy and $10-30×\times$ improvement in color harmony on GenColorBench benchmark.

\section*{Acknowledgments}
This work was supported by Grants PID2022-143257NB-I00, AIA2025-163919-C52, and PID2024-162555OB-I00 funded by MCIN/AEI/10.13039/501100011033 and the FEDER, by the Generalitat de Catalunya CERCA Program, by the grant Càtedra ENIA UAB-Cruïlla (TSI-100929-2023-
2) from the Ministry of Economic Affairs and Digital
Transition of Spain, and by the European Union’s Horizon Europe research and innovation programme under grant agreement number 101214398 (ELLIOT). JVC also acknowledges the 2025 Leonardo Grant for Scientific Research and Cultural Creation from the BBVA Foundation. The BBVA Foundation accepts no responsibility for the opinions, statements and contents included in the project and/or the results thereof, which are entirely the responsibility of the authors.
Kai Wang acknowledges the funding from Guangdong and Hong Kong Universities 1+1+1 Joint Research Collaboration Scheme and the start-up grant B01040000108 from CityU-DG. We acknowledge the EuroHPC Joint Undertaking for awarding us access to Leonardo at CINECA, Italy, and the RES resources provided by BSC on MareNostrum5 under project IM-2025-3-0025.


\bibliographystyle{splncs04}
\bibliography{longstrings,main}

\newpage
\setcounter{section}{0}
\setcounter{figure}{0}
\setcounter{table}{0}
\setcounter{equation}{0}

\renewcommand{\thesection}{S\arabic{section}}
\renewcommand{\thesubsection}{S\arabic{section}.\arabic{subsection}}

\renewcommand{\thefigure}{S\arabic{figure}}
\renewcommand{\thetable}{S\arabic{table}}
\renewcommand{\theequation}{S\arabic{equation}}

\begin{center}
{\Large \textbf{Supplementary Material}}
\end{center}

\section{Limitations in Baseline Methods}
\begin{figure}
\centering
\includegraphics[width=\textwidth]{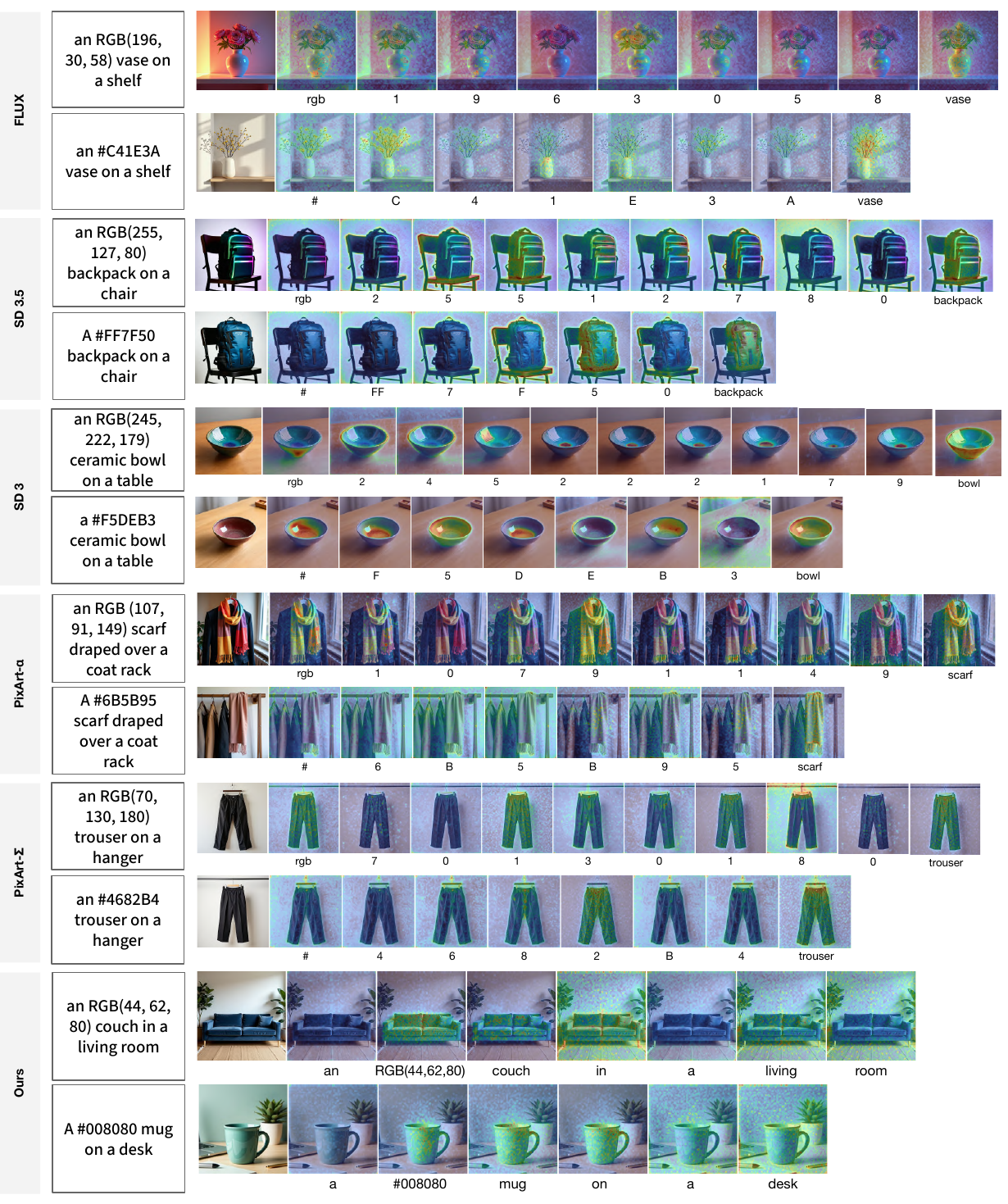}
\caption{Cross-attention visualization for numerical color prompts. Baseline models (FLUX, SD3.5, SD3, PixArt-$\alpha$, PixArt-$\Sigma$) distribute attention across fragmented color tokens (\texttt{rgb}, \texttt{1}, \texttt{9}, \texttt{6}, ...), failing to form coherent color representations. NumColor (bottom) aggregates color specifications into a single embedding that attends jointly to the target object, enabling accurate color control.}
\label{fig:attn_maps}
\end{figure}

Current text-to-image models fundamentally fail to interpret numerical color specifications due to tokenization fragmentation. When a hex code such as \texttt{\#C41E3A} or an RGB tuple like \texttt{RGB(196, 30, 58)} is passed to the text encoder, it is split into arbitrary subword tokens that bear no semantic relationship to the intended color. Figure~\ref{fig:attn_maps} visualizes this failure through cross-attention maps: baseline models distribute attention across individual tokens (\texttt{\#}, \texttt{C4}, \texttt{1E}, \texttt{3A}) rather than aggregating them into a coherent color representation. The resulting attention patterns show that each token fragment independently influences generation, producing rainbow artifacts, color bleeding, or complete disregard of the specified color. This fragmentation is consistent across all evaluated architectures---~\cite{flux2024}, SD3~\cite{stabilityai_sd3_medium_diffusers_2025}, SD3.5~\cite{stabilityai_sd3_5_large_2024}, PixArt-$\alpha$~\cite{chen2023pixartalpha}, and PixArt-$\Sigma$~\cite{chen2024pixartsigma}---indicating a fundamental limitation of subword tokenization for numerical color specifications rather than a model-specific failure.
 
\begin{figure}
\centering
\includegraphics[width=0.75\textwidth]{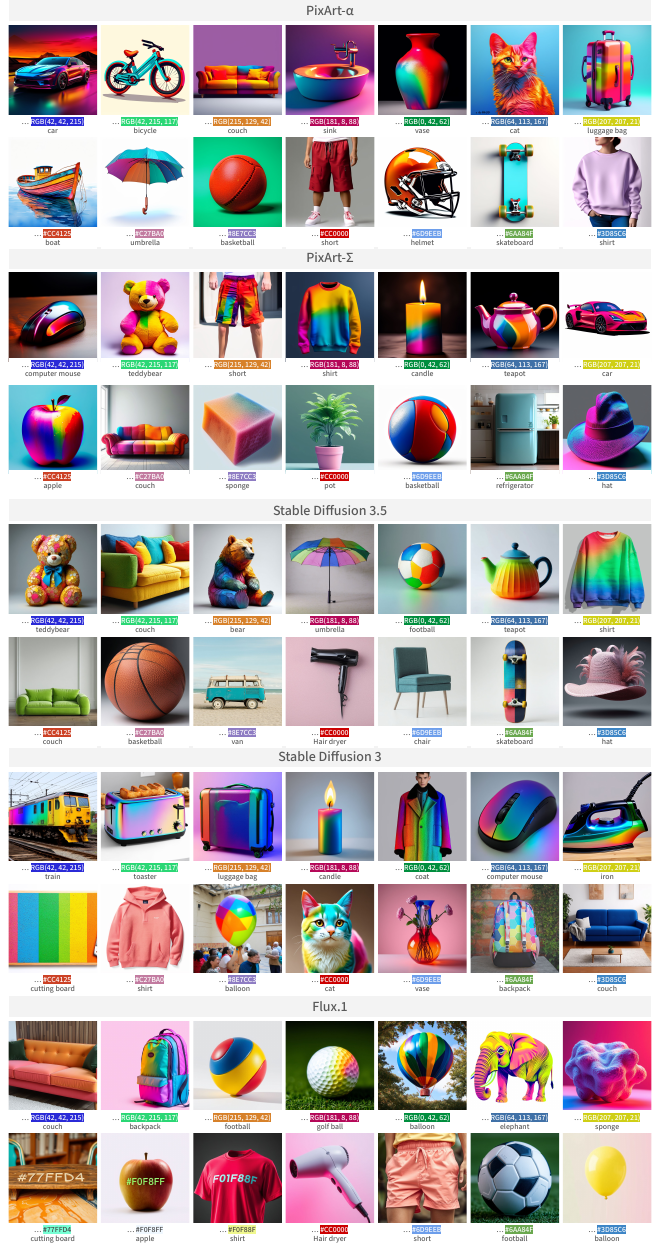}
\caption{Baseline model outputs for RGB and hex color specifications. Models fail to interpret numerical colors, producing rainbow artifacts, arbitrary colors, or complete disregard of the specification. This failure is consistent across architectures and color formats.}
\label{fig:baselines_hex_rgb}
\end{figure}

Figure~\ref{fig:baselines_hex_rgb} illustrates the qualitative consequences of this limitation. Baseline models produce outputs that ignore the specified RGB or hex values entirely, defaulting instead to arbitrary colors or rainbow gradients. In many cases, the numerical tokens appear to trigger psychedelic or multi-colored outputs, as the model interprets the fragmented digits as unrelated concepts rather than a unified color specification. NumColor addresses this by detecting color spans using Color Token Aggregator (CTA) and replacing the fragmented tokens with a single learned embedding from the ColorBook before the contextualization in text encoder. As shown in the bottom rows of Figure~\ref{fig:attn_maps}, NumColor produces coherent attention maps where the color embedding attends jointly to the target object, enabling accurate color-object binding.

\section{Category-wise Quantitative Results}
\begin{figure}[!t]
\centering
\includegraphics[width=\textwidth]{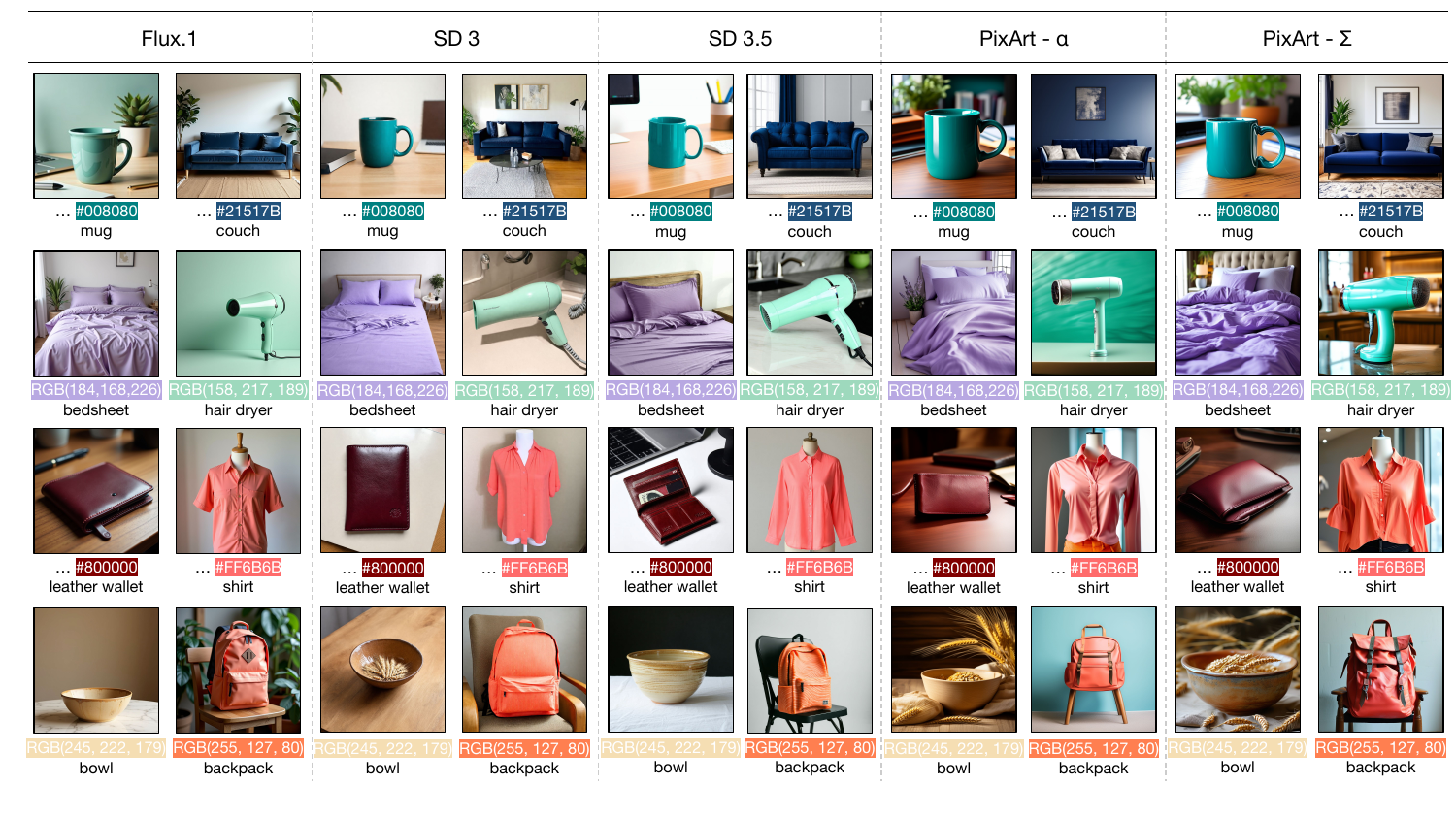}
\caption{Qualitative results of NumColor across five diffusion architectures. Each column shows outputs from a different model for identical prompts containing hex codes or RGB tuples. NumColor enables consistent numerical color control across all architectures using a single trained ColorBook, accurately rendering specified colors on diverse objects.}
\label{fig:t2i_gen}
\end{figure}

Our evaluation is based on the GenColorBench~\cite{butt2025gencolorbench} benchmark which provides 44K+ text prompts for five color evaluation tasks, spanning over 400+ colors and 100 objects pairs. The objects are grouped into seven semantic categories based on their color-object associations such as Clothing \& Accessories, Vehicles, Furniture \& Household, Tools \& Miscellaneous, Sports \& Toys, Animals, and Fruits \& Vegetables. In this paper, we use 9.8K prompts from the Numerical Color Understanding task of GenColorBench benchmark to evaluate the performance of NumColor. Table~\ref{tab:category_detailed} presents category-wise Numerical Color Understanding performance for NumColor across five diffusion architectures with NumColor. Clothing \& Accessories achieves highest accuracy across all configurations (62.1\% for FLUX on ISCC-L1), followed by Vehicles (59.4\%) and Furniture \& Household (56.8\%). Performance is notably lower for Animals (46.8\%) and Fruits \& Vegetables (36.2\%), where memory color bias interferes with color control---models trained on natural images encode strong priors that bananas are yellow, oranges are orange, and flamingos are pink, making it difficult to override these associations with user-specified colors. This performance gap between the categories is consistent across all models and color granularities, suggesting that while NumColor substantially improves numerical color understanding (4$\times$ over baselines), overcoming semantic priors for objects with strong color associations remains an open challenge. Some additional qualitative results are shown in Figure~\ref{fig:t2i_gen}.

\begin{table*}[t]
\centering
\caption{Detailed category-wise breakdown showing individual metric values (ISCC-L1 | ISCC-L3 | CSS3/X11) for each semantic category. The pattern reflects higher performance on stylistic color categories versus biologically intrinsic color categories.}
\label{tab:category_detailed}
\resizebox{\textwidth}{!}{
\begin{tabular}{lccccccc}
\toprule
\textbf{Model} & \textbf{Clothing \& Acc.} & \textbf{Vehicles} & \textbf{Furniture \& House.} & \textbf{Tools \& Misc.} & \textbf{Sports \& Toys} & \textbf{Animals} & \textbf{Fruits \& Veg.} \\
\midrule
\multicolumn{8}{c}{\textbf{ISCC-L1}} \\
\midrule
FLUX w/ NumColor & 62.1 & 59.4 & 56.8 & 54.2 & 53.6 & 46.8 & 36.2 \\
SD 3 w/ NumColor & 54.8 & 52.6 & 50.4 & 48.1 & 47.2 & 41.8 & 32.4 \\
SD3.5 w/ NumColor & 57.6 & 55.2 & 52.8 & 50.6 & 49.8 & 44.2 & 34.6 \\
PixArt-$\alpha$ w/ NumColor & 50.2 & 48.1 & 46.2 & 44.1 & 43.2 & 38.4 & 29.8 \\
PixArt-$\Sigma$ w/ NumColor & 54.2 & 52.1 & 50.2 & 48.1 & 47.2 & 41.8 & 32.6 \\
\midrule
\multicolumn{8}{c}{\textbf{ISCC-L3}} \\
\midrule
FLUX w/ NumColor & 53.2 & 50.8 & 48.6 & 46.1 & 45.8 & 40.2 & 31.4 \\
SD 3 w/ NumColor & 44.2 & 42.1 & 40.2 & 38.6 & 37.8 & 33.6 & 26.1 \\
SD3.5 w/ NumColor & 48.1 & 46.2 & 44.1 & 42.3 & 41.6 & 36.8 & 28.9 \\
PixArt-$\alpha$ w/ NumColor & 40.4 & 38.6 & 37.1 & 35.4 & 34.7 & 30.8 & 24.0 \\
PixArt-$\Sigma$ w/ NumColor & 45.6 & 43.8 & 42.1 & 40.4 & 39.6 & 35.1 & 27.4 \\
\midrule
\multicolumn{8}{c}{\textbf{CSS3/X11}} \\
\midrule
FLUX w/ NumColor & 57.8 & 55.2 & 52.4 & 50.8 & 50.6 & 44.1 & 35.2 \\
SD 3 w/ NumColor & 52.1 & 49.8 & 47.6 & 45.2 & 44.1 & 39.2 & 30.8 \\
SD3.5 w/ NumColor & 51.8 & 49.6 & 47.2 & 45.1 & 44.2 & 39.1 & 31.2 \\
PixArt-$\alpha$ w/ NumColor & 43.1 & 41.2 & 39.6 & 37.8 & 36.9 & 32.8 & 25.6 \\
PixArt-$\Sigma$ w/ NumColor & 48.8 & 46.6 & 44.8 & 42.9 & 42.1 & 37.2 & 29.1 \\
\bottomrule
\end{tabular}
}
\end{table*}

\section{Architecture and Implementation Details}
This section provides comprehensive architectural specifications and training details to enable full reproducibility of NumColor. We describe each component in detail, including design and implementation choices.

\subsection{Color Token Aggregator}
The Color Token Aggregator (CTA) addresses a fundamental challenge: tokenizers fragment numeric color representations including hex color codes and RGB triples. A hex code like \texttt{\#FF5733} may be split into \texttt{[\#, FF, 57, 33]} by one tokenizer and \texttt{[\#FF, 5733]} by another, making it impossible to design a single detection heuristic that works across models. Our proposed CTA is a learned sequence tagger that operates on character-level representations of tokens, achieving tokenizer-agnostic color detection. The architectural details are given in Table~\ref{tab:cta}.

\noindent\textbf{Character-level Token Encoding.} The encoder first converts each token into a fixed-dimensional embedding using a character-level convolutional neural network. Given a token represented as a sequence of character indices, we embed each character into a 64-dimensional space using a learned embedding table with vocabulary size 512, covering ASCII, extended Latin, and common Unicode characters. The character embeddings are then processed by three parallel 1D convolutional layers with kernel sizes $\{2, 3, 4\}$ to capture character n-grams of varying lengths. Each convolutional layer produces $\lfloor 256/3 \rfloor$ output channels, which are concatenated after max-pooling over the sequence dimension to get a 256-dimensional token embedding. This architecture allows the model to recognize patterns like '\texttt{FF}' or '\texttt{\#A}' regardless of how the tokenizer has segmented the input, since it operates on the raw characters within each token.

\noindent\textbf{Contextual Encoding with Transformers.} The token embeddings are augmented with sinusoidal positional encodings and passed through a 4-layer Transformer encoder. Each layer consists of multi-head self-attention with 4 heads and a feed-forward network with hidden dimension 512. The self-attention mechanism allows the model to incorporate context when making tagging decisions---for instance, recognizing that a token '\texttt{33}' following '\texttt{\#FF57}' is likely part of a hex code, whereas the same token in isolation (e.g., `33 degrees') is not. We apply dropout with probability 0.1 during training for regularization. The maximum sequence length is set to 256 tokens, which accommodates typical prompt lengths across target architectures.

\noindent\textbf{CRF decoding for structured prediction.} The final token representations are projected to a 3-dimensional tag space corresponding to tagging scheme: Begin (B), Inside (I), and Outside (O). Rather than making independent predictions for each token, we employ a Conditional Random Field (CRF) layer that models transition probabilities between adjacent tags. The CRF is initialized with hard constraints that enforce valid sequences: transitions from O to I are assigned a score of $-10000$, effectively prohibiting I tags that are not preceded by B tags. Similarly, starting a sequence with I is disallowed. During training, we minimize the negative log-likelihood of the tag sequence under the CRF model. At inference, we use the Viterbi algorithm to find the most likely tag sequence in $O(T \cdot K^2)$ time, where $T$ is the sequence length and $K=3$ is the number of tags.

\begin{table}[!t]
\centering
\caption{Color Token Aggregator (CTA) Architecture.}
\label{tab:cta}
\small
\begin{tabular}{ll}
\toprule
\textbf{Component} & \textbf{Configuration} \\
\midrule
\multicolumn{2}{l}{\textit{Character-Level CNN}} \\
\quad Char embedding dim & 64 \\
\quad Char vocab size & 512 \\
\quad CNN kernel sizes & \{2, 3, 4\} \\
\quad Token embedding dim & 256 \\
\quad Max chars per token & 32 \\
\midrule
\multicolumn{2}{l}{\textit{Transformer Encoder}} \\
\quad Layers & 4 \\
\quad Attention heads & 4 \\
\quad Hidden dim & 256 \\
\quad FFN dim & 512 \\
\quad Max sequence length & 256 \\
\midrule
\multicolumn{2}{l}{\textit{CRF Decoder}} \\
\quad Num tags & 3 (B, I, O) \\
\quad Constrained transitions & \checkmark \\
\midrule
\multicolumn{2}{l}{\textit{Regularization}} \\
\quad Dropout & 0.1 \\
\quad Positional encoding & Sinusoidal \\
\midrule
\textbf{Total parameters} & $\sim$1.2M \\
\bottomrule
\end{tabular}
\end{table}

\subsection{ColorBook}
The ColorBook is the core trainable component of NumColor, responsible for mapping continuous color coordinates to the 4096-dimensional embedding space of the text encoder. Unlike approaches that attempt to teach the text encoder new color tokens, our ColorBook operates as a lookup table with soft interpolation, enabling precise control over arbitrary colors without modifying the frozen text encoder.

\noindent\textbf{Anchor grid construction.} We construct a discrete set of anchor colors by sampling the CIE Lab color space on a regular grid with approximately 5-unit spacing along each axis. The Lab color space is chosen for its perceptual uniformity. For instance, a Euclidean distance of $\Delta E$ units in Lab space corresponds roughly to the same perceived color difference regardless of the base color, unlike RGB or HSV spaces where perceptual uniformity varies significantly across the gamut. We filter the grid to retain only colors that fall within the sRGB gamut, resulting in 6,707 anchor colors. Each anchor is stored as a 3-dimensional Lab coordinate $\mathbf{a}_i \in \mathbb{R}^3$ and associated with a learnable 4096-dimensional embedding $\mathbf{e}_i \in \mathbb{R}^{4096}$.

\noindent\textbf{Initialization strategy.} A naive initialization strategy would be to associate each anchor with an arbitrary named color and initialize the embedding from the T5 text encoder representation of that name. However, many color names carry semantic contamination: text embedding for ‘eggplant’ carries associations with vegetables, not just the color purple; ‘salmon’ evokes fish; ‘chocolate’ evokes food. These spurious associations can leak into generated images, causing unwanted content to appear when users specify colors that happen to map to semantically-loaded names.

We address this through careful curation of initialization vocabulary, drawing from two complementary sources. First, we use the ISCC-NBS color naming system~\cite{kelly1976color}, specifically the 12 Level 1 category names: red, orange, yellow, green, blue, purple, pink, brown, olive, white, gray, and black. These basic color terms are maximally generic and carry minimal non-color associations. Second, we incorporate the CSS3/X11 extended color set, which provides 140+ standardized color names spanning the full gamut. Critically, we filter a set to exclude names with strong non-color semantics---we retain 'crimson','turquoise', and 'lavender' whose primary meaning is the color, but exclude 'tomato','chocolate', 'salmon', and similar names where the color sense is secondary to a concrete object. For each anchor color, we find its nearest named color from this curated vocabulary based on Lab distance. The anchor's embedding is then initialized as the T5 embedding of that color name. For multi-word names (e.g., ‘light blue’,’dark green’), we average the embeddings of constituent tokens. This hybrid strategy provides dense coverage of the color space with named colors available at finer granularity than the 12 ISCC-NBS categories alone while avoiding semantic contamination from object-associated color names.

\noindent\textbf{Soft nearest-neighbor interpolation.} Given an input Lab coordinate $\mathbf{c}$, the ColorBook computes an output embedding via soft interpolation over the $k$ nearest anchor colors. We first compute Euclidean distances to all anchors and select the top-$k$ nearest neighbors $\mathcal{N}_k(\mathbf{c}) = \{i_1, \ldots, i_k\}$. The output embedding is then computed as a weighted average:
\begin{equation}
\mathbf{e}(\mathbf{c}) = \sum_{j=1}^{k} w_j \cdot \mathbf{e}_{i_j}, \quad \text{where} \quad w_j = \frac{\exp(-d_{i_j} / \tau)}{\sum_{\ell=1}^{k} \exp(-d_{i_\ell} / \tau)}
\end{equation}
and $d_i = \|\mathbf{c} - \mathbf{a}_i\|_2$ is the Lab distance to anchor $i$. The temperature parameter $\tau$ controls the sharpness of the interpolation: lower temperatures concentrate weight on the nearest anchor, while higher temperatures distribute weight more evenly. We use $\tau = 2.0$ based on ablation studies, which achieves a balance between precise color reproduction and smooth transitions between nearby colors.

The top-$k$ restriction serves both computational and statistical purposes. Computationally, it reduces the softmax from 6,707 terms to $k=8$ terms. Statistically, it prevents distant anchors from contributing noise to the interpolation: even with a softmax, very distant anchors would receive non-zero weight that accumulates across thousands of anchors. With $k=8$, the 8 nearest anchors capture colors within approximately 10-15 $\Delta E$ units, ensuring all contributing anchors are perceptually similar to the query color.



\subsection{Pre-Contextualization Embedding Injection}
A critical architectural decision in NumColor is \textit{where} to inject color embeddings into the text encoder pipeline. We inject embeddings before T5 self-attention layers, referred to as pre-contextualization, not after the encoder has already produced contextualized representations. This distinction is essential for proper color-object binding.

\noindent\textbf{Why pre-contextualization matters.} In Transformer-based text encoders, semantic relationships between tokens are established through self-attention. When processing ‘a red car’, the attention mechanism allows ‘car’ to attend to ‘red’, creating a joint representation that associates the color with the object. If we inject color embeddings \textit{after} the encoder forward pass, the self-attention has already been computed without knowledge of the color---the ‘car’ representation was contextualized with whatever fragmented tokens the hex code produced, not with a semantically meaningful color embedding. Post-contextualization injection is therefore architecturally broken for color-object binding. Pre-contextualization injection ensures that when the Transformer processes the sequence, the color embedding participates fully in self-attention. Object tokens can attend to the color embedding, and the color embedding can attend to object tokens, establishing the bidirectional associations necessary for accurate color control.







\noindent\textbf{Handling multiple colors.} When a prompt contains multiple hex codes, we process them in reverse order (right-to-left) to ensure that position indices remain valid as spans are collapsed. Each color is injected independently, and the sequence is compacted after each injection. This allows prompts like `'a \texttt{\#FF0000} car next to a \texttt{\#00FF00} tree' to correctly associate each color with its respective object.

\subsection{Training Configuration}

NumColor training updates only the ColorBook embeddings; all other components including CTA, T5, CLIP, DiT/Transformer, and VAE remain frozen. This design minimizes the risk of catastrophic forgetting and ensures that the capabilities of the base model are preserved.

\noindent\textbf{Optimization.} We use the AdamW~\cite{loshchilov2018decoupled} optimizer with learning rate $1 \times 10^{-4}$, weight decay 0.01, and $(\beta_1, \beta_2) = (0.9, 0.999)$. The learning rate follows a cosine annealing schedule with 1,000 warmup steps, decaying to $1 \times 10^{-6}$ by the end of training. We train for 100K steps with an effective batch size of 16 (batch size 1 per GPU with gradient accumulation over 4 steps across 4 GPUs). Training is conducted in bfloat16 precision to reduce memory consumption and accelerate computation.





\noindent\textbf{Training Setup.} Training is conducted on 4 NVIDIA A100 GPUs (64GB memory) using PyTorch Distributed Data Parallel (DDP) with the NCCL backend. We use torchrun for multi-GPU coordination. Total training time is approximately 136 hours for FLUX.1-dev.

Table~\ref{tab:training} summarizes the training configuration.

\begin{table}[!t]
\centering
\caption{Training configuration for ColorBook training. Only ColorBook embeddings are updated; all other model components remain frozen throughout training.}
\label{tab:training}
\small
\begin{tabular}{ll}
\toprule
\textbf{Hyperparameter} & \textbf{Value} \\
\midrule
\multicolumn{2}{l}{\textit{Optimization}} \\
\quad Optimizer & AdamW \\
\quad Learning rate & $1 \times 10^{-4}$ \\
\quad Learning rate schedule & Cosine annealing \\
\quad Warmup steps & 1,000 \\
\quad Final learning rate & $1 \times 10^{-6}$ \\
\quad Weight decay & 0.01 \\
\quad Adam $\beta_1, \beta_2$ & 0.9, 0.999 \\
\midrule
\multicolumn{2}{l}{\textit{Batch Configuration}} \\
\quad Batch size per GPU & 1 \\
\quad Gradient accumulation steps & 4 \\
\quad Number of GPUs & 4 \\
\quad Effective batch size & 16 \\
\quad Total training steps & 100,000 \\
\midrule
\multicolumn{2}{l}{\textit{Loss Weights}} \\
\quad Flow matching & 1.0 \\
\quad Directional alignment $\lambda_\text{d}$ & 0.3 \\
\quad Interpolation consistency $\lambda_\text{i}$ & 0.2 \\
\midrule
\multicolumn{2}{l}{\textit{Frozen Components}} \\
\quad Color Token Aggregator & \checkmark (pretrained separately) \\
\quad T5-XXL text encoder & \checkmark \\
\quad CLIP text encoder(s) & \checkmark \\
\quad DiT / Transformer backbone & \checkmark \\
\quad VAE encoder/decoder & \checkmark \\
\midrule
\multicolumn{2}{l}{\textit{Precision and Infrastructure}} \\
\quad Training precision & bfloat16 \\
\quad GPUs & 4$\times$ NVIDIA A100 (64GB) \\
\quad Distributed strategy & DDP (torchrun + NCCL) \\
\quad Approximate training time & 136 hours \\
\bottomrule
\end{tabular}
\end{table}

\subsection{Cross-Architecture Deployment}
A key advantage of NumColor is its applicability across multiple diffusion architectures without retraining. The ColorBook operates on T5 text embeddings which are shared across FLUX, Stable Diffusion 3, and PixArt family. Therefore, a single trained NumColor can be deployed to all these models. The only architecture-specific consideration is the injection point, which is always the T5 \texttt{embed\_tokens} output, before encoder layers.

\noindent\textbf{FLUX.1-dev.} FLUX uses a joint T5-XXL and CLIP-L text encoding scheme. T5 provides the primary semantic conditioning with 4096-dimensional embeddings, while CLIP-L provides a 768-dimensional pooled embedding for global conditioning. NumColor injects color embeddings only into the T5 branch; the CLIP branch processes the original prompt text and provides complementary scene-level information. The default inference configuration uses 28 denoising steps with classifier-free guidance scale 3.5 at 1024$\times$1024 resolution.

\noindent\textbf{Stable Diffusion 3 and 3.5.} SD3 variants use T5-XXL combined with CLIP-G (OpenCLIP ViT-bigG). The T5 branch receives color-injected embeddings via NumColor, while CLIP-G processes the original prompt. SD3 uses a 77-token context for CLIP and up to 256 tokens for T5. We use 28 steps with guidance scale 7.0 for SD3 and 4.5 for SD3.5 Medium, following recommended baseline settings.

\noindent\textbf{PixArt-$\alpha$ and PixArt-$\Sigma$.} PixArt models use T5-XXL as the sole text encoder, without a CLIP branch. This makes NumColor integration straightforward---all conditioning flows through the color-injected T5 embeddings. PixArt-$\alpha$ supports 120-token sequences, while PixArt-$\Sigma$ extends this to 300 tokens. Both use 20 denoising steps with guidance scale 4.5.

\subsection{Inference Overhead}
NumColor introduces minimal computational overhead at inference time. The CTA processes the tokenized prompt in a single forward pass, requiring less than 1 millisecond on A100 GPU. The ColorBook lookup involves computing distances to 6,707 anchors and performing a top-8 softmax, which completes in under 0.1 milliseconds per color. For a typical prompt with 1--3 colors, the total NumColor overhead is under 2 milliseconds which is negligible compared to the 5--30 seconds required for diffusion sampling.

Overall, the memory overhead consists of the 27.5M ColorBook parameters and 1.2M CTA parameters, totaling 28.7M additional parameters. In float32 precision, this corresponds to approximately 115MB of GPU memory; in bfloat16/float16, approximately 58MB. This is a small fraction of the base model memory footprint.

\section{Dataset and Evaluation}
This section details the data used for training and evaluation, as well as our evaluation methodology for measuring color accuracy in generated images.

\subsection{Training Data for CTA}
The CTA must learn to recognize hex color codes and RGB triplets across diverse tokenization patterns produced by different text encoders. To achieve this, we curate 50K base prompts from DiffusionDB~\cite{wang2023diffusiondb} that originally contain color name mentions, and systematically replace these color names with hex codes or RGB tuples. This approach yields naturalistic prompt structures while introducing numerical color specifications.

The prompt distribution is balanced across color complexity: 20\% contain no colors (negative examples), 20\% contain one color, 20\% contain two colors, 20\% contain three colors, and 20\% contain four colors. This balanced distribution ensures the model learns to handle both simple single-color prompts and complex multi-color compositions. Color formats are split evenly between hex codes (e.g., \texttt{\#FF0000}) and RGB tuples (e.g., \texttt{RGB(255,0,0)}), training the encoder to recognize both numerical representations. To expose the model to the full range of tokenization behaviors encountered in practice, we tokenize each prompt using four different text encoders: T5 (SentencePiece), CLIP-L (BPE), CLIP-G (BPE with different vocabulary), and GPT-2 (BPE). This accumulates to 200K total training samples (50K prompts $\times$ 4 tokenizers). The diverse tokenization patterns are critical. For example, the prompt 'a \texttt{\#FF0000} car' tokenizes as \texttt{[\_a, \_\#, FF, 00, 00, \_car]} under T5 but as \texttt{[a</w>, \#, ff, 00, 00</w>, car</w>]} under CLIP, and the encoder must learn to detect color spans regardless of these variations. We split the data into 90\% training and 10\% validation, ensuring that specific hex codes appearing in validation do not appear in training to test generalization to unseen colors.

\subsection{Training Data for ColorBook}
\begin{figure*}[t]
    \centering
    \includegraphics[width=\linewidth]{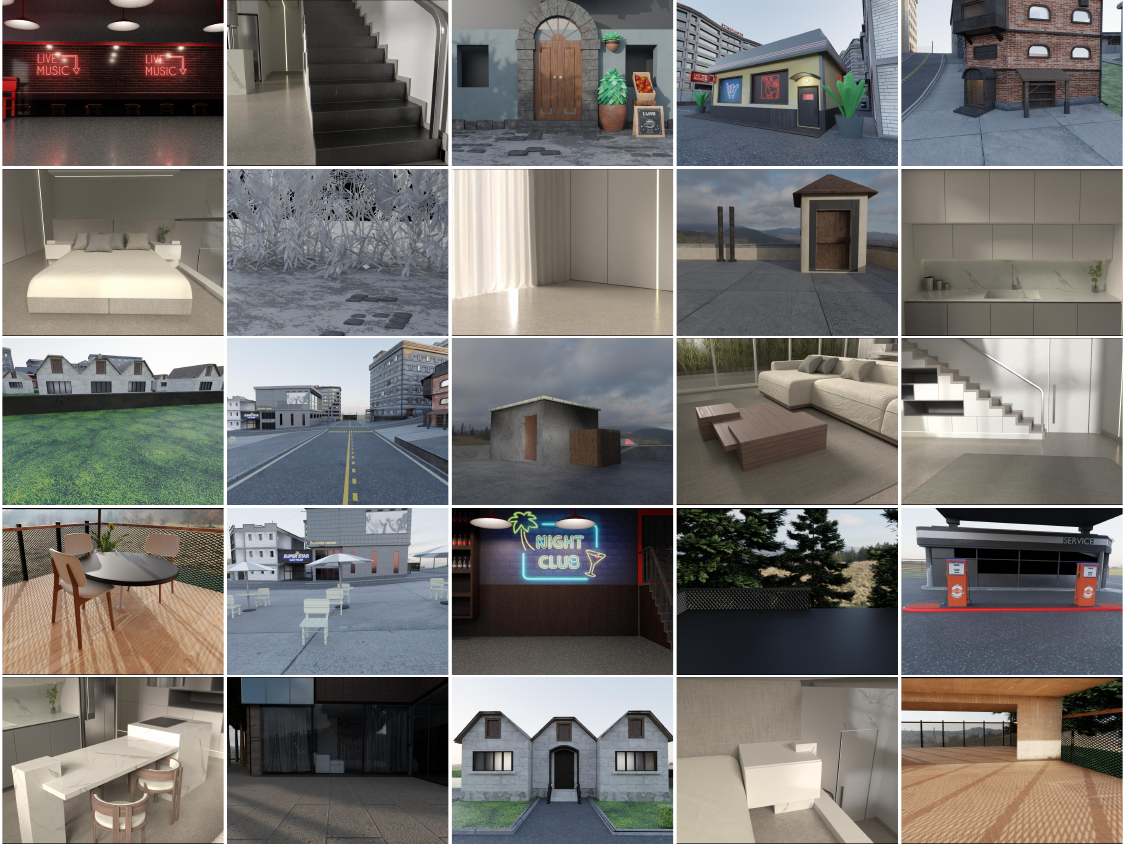}
    \caption{Blender-rendered scenes used for sample objects for training data generation. We render diverse indoor and outdoor environments spanning residential interiors, urban streetscapes, architectural exteriors, commercial spaces, and natural settings. Colored objects are composited into these scenes with randomized placement, lighting, and camera viewpoints to generate training pairs for the NumColor ColorBook.}
    \label{fig:dataset_scenes}
\end{figure*}

\begin{figure}[!t]
\centering
\includegraphics[width=\textwidth]{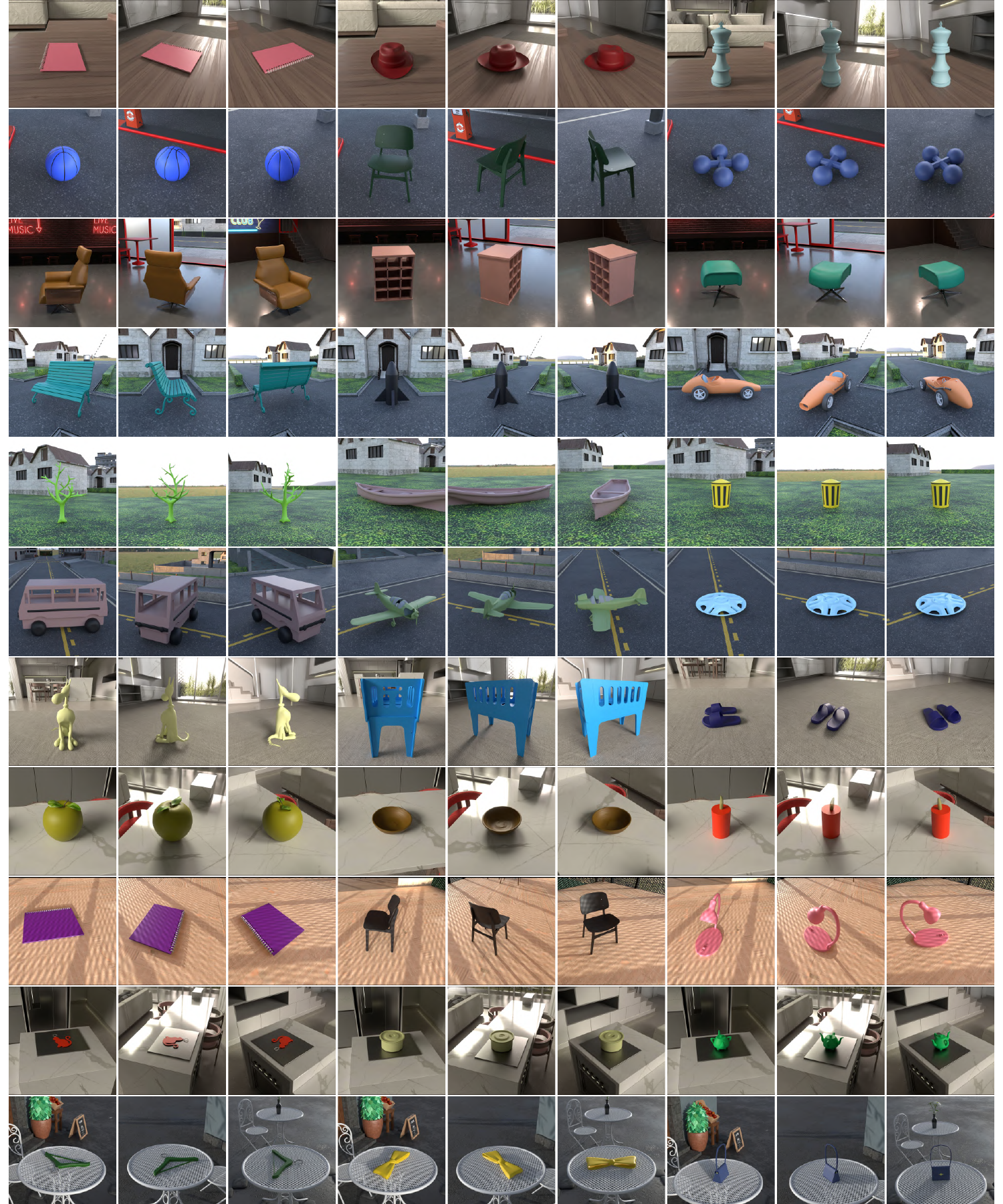}
\caption{Demonstrating Samples from NumColor-Data.}
\label{fig:dataset_samples}
\end{figure}

The ColorBook is trained end-to-end with the frozen diffusion model using synthetic prompts that pairs numeric colors with objects. Since, there is no precise color dataset exists, we curate a set of opensource 3D scenes including 30 scenes as shown in Figure~\ref{fig:dataset_scenes}, and 100 objects from Objaverse~\cite{deitke2023objaverse} dataset. These objects are rendered in these scenes in 6707 colors from the ColorBook, as shown in Figure~\ref{fig:dataset_samples}. For each scene, one each is passed to GPT to get text descriptions, which is validated by the human experts. Then, the final template prompts are curated to pair with the rendered images. This structure yields prompts such as 'a \texttt{\#3A7BD5} ceramic vase on a wooden shelf'. During training, we continuously sample with random hex codes and RGB triplets to provide effectively color variation.

\subsection{Evaluation Data: GenColorBench}
We evaluate NumColor using the Numerical Color Understanding (NCU) task from GenColorBench~\cite{butt2025gencolorbench}, a comprehensive benchmark designed to assess color generation capabilities of text-to-image models. GenColorBench provides 44K+ prompts covering 400+ colors across five evaluation tasks; we focus specifically on the NCU task, which tests whether models can correctly render colors specified as hex codes or RGB values which is a precise capability NumColor is designed to enable in T2I generation models.

\noindent\textbf{Prompt structure.}
The NCU evaluation set contains object-focused prompts that specify colors numerically rather than by name. Each prompt describes a single colored object with either a hex code (e.g., ``a \texttt{\#FF5733} apple'') or an RGB tuple (e.g., ``a \texttt{RGB(255,87,51)} apple''). This controlled setting isolates the core capability we aim to evaluate: whether the model can accurately render a specific numerical color on a target object, independent of compositional complexity or scene context. By focusing on single-object prompts, we obtain a clean signal of color accuracy without confounds from multi-object binding errors or color leakage between scene elements. Following prompts templates are used to generate object-focused prompts.

\begin{enumerate}
    \item A \{color\} \{object\}
    \item The \{object\} is \{color\}
    \item A photo of a \{color\} \{object\}
    \item A \{object\} that is entirely \{color\}
    \item An image of a \{color\} \{object\}
    \item A \{color\} colored \{object\}
    \item A single \{color\} \{object\}
    \item A \{object\}, and it’s \{color\}
    \item A \{object\} in a \{color\} color
    \item A \{object\} rendered in \{color\} color
    \item A \{object\} with a \{color\} color
    \item A realistic \{object\} in \{color\}
    \item An image of a \{object\} in hex color \{hex\}
    \item A \{object\} in color \{hex\}
    \item A \{object\} with hex color \{hex\}
    \item A close-up of a \{object\} in the color \{hex\}
    \item A \{object\} rendered in \{hex\} color
    \item A photo of a \{object\} in the color \{hex\}
    \item A \{object\} rendered entirely in \{hex\}
    \item A \{object\} designed in \{hex\} color
    \item A realistic \{hex\}-colored \{object\}
    \item A highly detailed \{object\} in hex \{hex\}
    \item A \{object\} in rgb(\{r\}, \{g\}, \{b\})
    \item A \{object\} with the color rgb(\{r\}, \{g\}, \{b\})
    \item A \{object\} rendered in RGB color rgb(\{r\}, \{g\}, \{b\})
    \item A photo of a \{object\} in color rgb(\{r\}, \{g\}, \{b\})
    \item A \{object\} with color rgb(\{r\}, \{g\}, \{b\})
\end{enumerate}

\noindent\textbf{Color Coverage.} The benchmark draws colors from two complementary color systems. The ISCC-NBS~\cite{kelly1976color} provides perceptually organized colors at three granularity levels: Level 1 (13 basic categories), Level 2 (29 intermediate hues), and Level 3 (267 fine-grained distinctions). The CSS3/X11 specification provides 147 web-standard colors with precise RGB definitions. Together, these ensure evaluation spans the full gamut with both perceptually meaningful categories and numerically precise targets.

\noindent\textbf{Object Categories.}
Evaluation objects are drawn from COCO~\cite{lin2014microsoft} and ImageNet~\cite{deng2009imagenet}, grouped into seven semantic categories: fruits and vegetables, tools and miscellaneous items, vehicles, animals, clothing and accessories, furniture and household objects, and sports and toys. This categorization enables analysis of how color accuracy varies with object semantics---particularly important given that some objects have strong canonical color associations (yellow bananas, red fire trucks) that may bias generation. 

\subsection{Embedding Drift Analysis}
\begin{figure}[t]
\centering
\includegraphics[width=\textwidth]{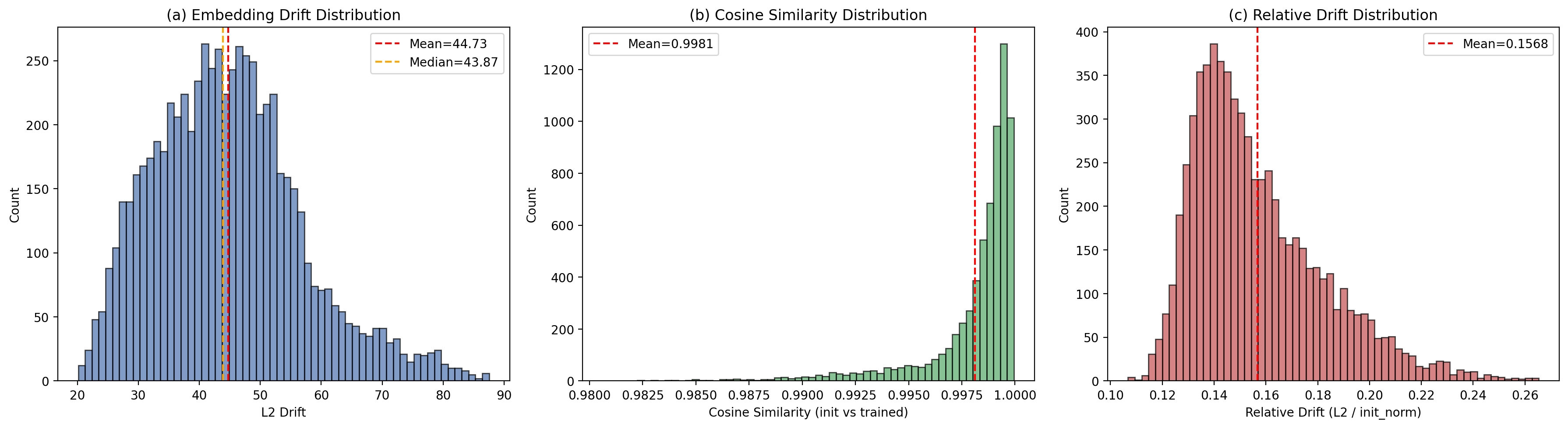}
\caption{Embedding drift statistics comparing initialized and trained ColorBook embeddings. (a) L2 drift distribution showing substantial movement. (b) Cosine similarity distribution showing directional preservation. (c) Relative drift distribution normalized by embedding magnitude. Together, these statistics indicate that training reorganizes the embedding space substantially while preserving directional compatibility with T5 geometry.}
\label{fig:drift_stats}
\end{figure}

We compute drift statistics comparing initialized and trained embeddings for each anchor to quantify how much embeddings change during training. Figure~\ref{fig:drift_stats} presents three complementary measures.

\paragraph{Absolute drift.} The L2 distance between initialized and trained embeddings measures absolute movement in the embedding space. Across all 6,707 anchors, the mean L2 drift is 44.73 (median: 43.87), with a roughly symmetric distribution spanning from approximately 20 to 90. This substantial drift indicates that training does not merely fine-tune embeddings with small perturbations which reorganizes the embedding space to better capture color relationships. For context, the mean L2 norm of initialized embeddings is approximately 280, so a drift of 45 represents meaningful movement.

\paragraph{Directional preservation.} Despite large absolute drift, the cosine similarity between initialized and trained embeddings remains extremely high: mean 0.998 across all anchors, with most values concentrated above 0.995. This indicates that while embeddings move substantially in magnitude and position, they largely preserve their directional relationships within the high-dimensional space. The training process adjusts \textit{where} embeddings are positioned relative to each other while maintaining compatibility with the broader T5 embedding geometry. This directional stability likely contributes to the ability of NumColor to integrate seamlessly with the frozen text encoder.

\paragraph{Relative drift.} We compute relative drift as the L2 distance divided by the initialized embedding norm to normalize for varying embedding magnitudes. The mean relative drift is 15.68\%, indicating that embeddings move approximately one-sixth of their original magnitude on average. This relative measure confirms that drift is substantial but bounded which means that embeddings reorganize meaningfully without collapsing or exploding.

\subsection{Discussion}
The combination of high absolute drift with preserved directional similarity reveals how NumColor transforms coarse color representations into precise ones. Pretrained T5 embeddings encode color at a categorical level---red, blue, and green occupy distinct regions, but the continuous gradations between them (coral, teal, chartreuse) are not systematically organized. T5 was trained on language modeling objectives where the distinction between ‘light red’ and ‘dark red’ matters far less than the distinction between red and blue. As a result, perceptually similar colors may be scattered across the embedding space, while perceptually distant colors with related names (e.g., sea green and sea blue) may cluster together due to shared semantic content.

NumColor training transforms this coarse categorical structure into a precise perceptual representation. The ColorBook learns to \textit{rearrange} embeddings within the existing T5 space such that Euclidean proximity in the embedding space corresponds to perceptual proximity in Lab space. The high cosine similarity indicates that this reorganization preserves compatibility with the frozen T5 encoder where embeddings move substantially in absolute terms but maintain their directional relationships within the broader embedding geometry. The training objective drives this refinement through end-to-end supervision. These findings validate our architectural choices. The soft interpolation mechanism (top-$k$ nearest neighbors with temperature-scaled softmax) requires that nearby anchors produce similar embeddings; the learned geometry satisfies this by placing perceptually similar colors nearby. The ISCC-NBS and CSS3/X11 initialization provides categorical color structure that training refines into continuous perceptual structure, rather than learning from scratch. And freezing all components except the ColorBook focuses optimization on color precision without disrupting T5 broader linguistic capabilities or the diffusion model's generation quality.

\end{document}